\documentclass[10pt,twocolumn,letterpaper]{article}
\newcommand{\incre}[1]{\textcolor{teal!90}{#1}}
\usepackage{cvpr}              
\usepackage{multirow}
\usepackage{makecell}
\usepackage{float}

\usepackage{colortbl}
\usepackage{array}
\usepackage{amsmath}
\usepackage{multirow}
\usepackage{makecell}
\usepackage{graphicx}
\usepackage{algorithmicx,algorithm}
\usepackage[dvipsnames,svgnames]{xcolor}
\usepackage{svg}
\usepackage{alltt}

\usepackage{caption}
\usepackage{wrapfig}
\definecolor{cvprblue}{rgb}{0.21,0.49,0.74}
\usepackage[pagebackref,breaklinks,colorlinks,allcolors=cvprblue]{hyperref}

\def\paperID{2822} 
\def\confName{CVPR}
\def\confYear{2026}

\title{AnchorOPT: Towards Optimizing Dynamic Anchors for \\ Adaptive Prompt Learning}

\author{
Zheng Li\textsuperscript{\rm 1},
Yibing Song,
Xin Zhang\textsuperscript{\rm 1}, 
Lei Luo\textsuperscript{\rm 2}
Xiang Li\textsuperscript{\rm 1}\thanks{Corresponding author.},
Jian Yang\textsuperscript{\rm 1}\footnotemark[1]\\
\textsuperscript{\rm 1} Nankai University,
\textsuperscript{\rm 2} Nanjing University of Science and Technology \\
{\tt\small zhengli97@foxmail.com,}
{\tt\small \{xiang.li.implus, csjyang\}@nankai.edu.cn} \\
}

\begin{document}
\maketitle

\begin{abstract}
    
    Existing prompt learning methods, which are built upon CLIP models, leverage textual tokens as anchors to guide the learnable soft tokens. This guidance improves CLIP generalizations. However, these anchors—static in both value and position—lack cross-task and stage-adaptive flexibility. To address this limitation, we propose AnchorOPT, a dynamic anchor-based prompt learning framework. Specifically, AnchorOPT introduces dynamism in two key dimensions: (i) anchor values eschew handcrafted explicit textual tokens (e.g., ``shape", ``color"), instead learning dynamically from task-specific data; and (ii) the positional relationship between anchor and soft tokens is no longer fixed but adaptively optimized via a learnable position matrix conditioned on the training stage and task context. Training occurs in two stages: we first learn the anchor tokens, then freeze and transfer them to the second stage for optimization of soft tokens and the position matrix. Extensive experiments demonstrate that using only a simple learnable anchor and position matrix achieves performance comparable to or exceeding some methods incorporating additional learnable modules or regularization techniques. As a plug-and-play module, AnchorOPT integrates seamlessly into existing frameworks, yielding consistent performance gains across diverse datasets. Code is publicly available at \url{https://github.com/zhengli97/ATPrompt}.
    
    
\end{abstract}

\section{Introduction}
\begin{figure*}[t]
    \centering
    \includegraphics[width=0.95\linewidth]{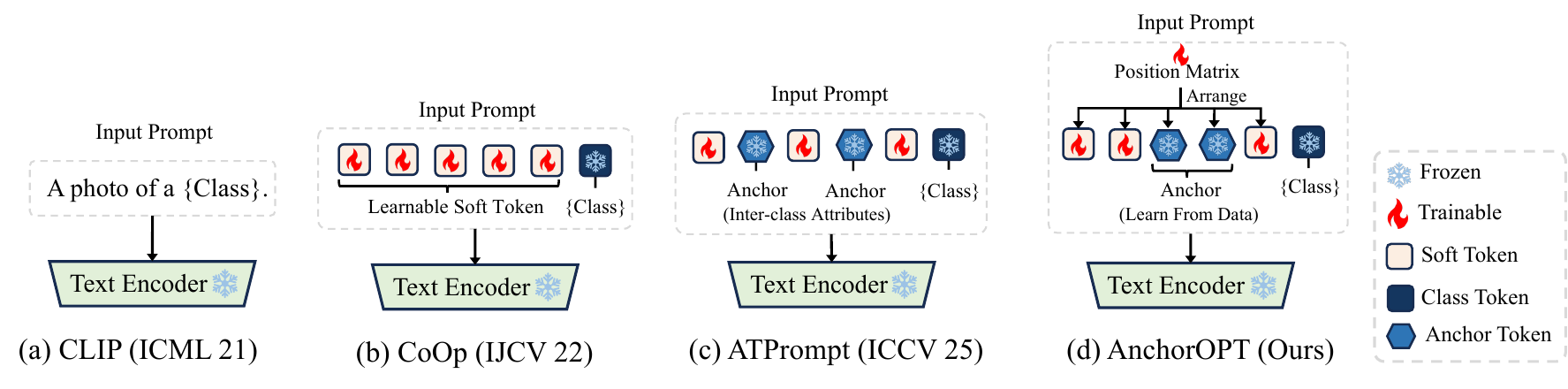}
    \vspace{-10pt}
    \caption{Comparison of fundamental input prompt structures. (a) CLIP\cite{radford2021learning} employs manually designed text templates. (b) CoOp\cite{zhou2022learning} introduced prompt learning for CLIP adaptation using learnable soft tokens concatenated with fixed category tokens. (c) ATPrompt~\cite{li2025advancing} incorporates explicit, fixed attribute tokens~(e.g., ``color", ``shape") to guide soft token learning via an attribute-based template. (d) AnchorOPT utilizes implicit anchors learned from data to guide the learning of soft tokens and proposes a learnable position matrix that dynamically adjusts the prompt sequence according to downstream requirements.
}
\label{fig:compare}
\vspace{-10pt}
\end{figure*}

Vision-Language Models (VLMs)~\cite{tan2019lxmert,jia2021scaling,radford2021learning,sun2024alpha}, such as CLIP~\cite{radford2021learning} and ALIGN~\cite{jia2021scaling}, trained on large-scale web data, have achieved state-of-the-art zero-shot performance on downstream tasks. These models are pre-trained using a contrastive loss to align image and text modalities into a unified embedding space, enabling cross-modal semantic alignment. To further enhance the utilization of pre-trained VLMs, prompt learning~\cite{zhou2022learning} has emerged as a critical methodology for adapting existing models to downstream tasks through the fine-tuning of supplemental soft tokens. CoOp~\cite{zhou2022learning} first proposes to replace the fixed handcrafted template~(e.g., \textit{``A photo of a \{Class\}."}) with multiple continuous soft tokens as the input of the text encoder, as shown in Fig.~\ref{fig:compare}(b). Subsequent research efforts have attempted to extend textual prompt learning to the vision~\cite{bahng2022exploring,pei2024sa2vp} and multimodal~\cite{khattak2023maple,li2024promptkd} domains, aiming to enhance the coherence of representations across modalities. 

Existing mainstream textual-based prompt learning methods~\cite{zhou2022learning,zhou2022conditional,khattak2023maple,li2024promptkd,xie2025textrefiner,li2025dpc} primarily adopt an approach that combines multiple learnable soft tokens with fixed class tokens for optimization. However, this paradigm restricts soft tokens to learning only category-specific information derived from training data, thereby limiting their capacity to capture generalizable, category-invariant representations~\cite{yao2023visual,zhu2023prompt,khattak2023self}. To address this limitation, recent studies~\cite{kan2023knowledge,ren2023prompt,tian2024argue} have introduced supplementary information into existing prompts. A common strategy~\cite{chen2023ovarnet,tian2024argue,zheng2024large,li2025advancing} involves inserting key tokens as anchors into input prompts to guide soft tokens toward learning domain-invariant general representations and bridging the gap between base and novel classes~\cite{li2025advancing}. These anchor tokens are typically derived from explicit class attributes~\cite{tian2024argue,li2025advancing,chen2023ovarnet}.
For instance, 
ATPrompt~\cite{li2025advancing} uses the attributes mined by NAS as anchors to form a novel attribute-based prompt learning template. It can be integrated as a plug-in module in existing methods and has achieved extensive improvements.

However, these textual anchors face two key limitations: manual content curation and positional rigidity, as illustrated in Fig.~\ref{fig:compare}(c). First, anchor tokens are typically manually defined with explicit textual content, embedding human priors that may conflict with domain-specific requirements. Second, while humans dynamically optimize sentence structures to suit contextual demands, existing methods rigidly fix the token positions within prompts, neglecting the need for structural adaptation across tasks.

To address these limitations, we propose AnchorOPT, a dynamic anchor-based prompt learning framework for CLIP. It features two innovations:
(1) Dynamic anchor values: Explicit static anchors are converted into implicit, learnable tokens that acquire universal cross-category representations directly from data rather than manual definition.
(2) Dynamic positional configuration: A learnable position matrix adaptively reconfigures token positions via Gumbel-Softmax~\cite{jang2016categorical} optimization, conditioned on task context.

Specifically, our training consists of two distinct stages: the anchor optimization stage and the adaptation stage. 
In the first stage, we propose a customized anchor learning template: \textit{``\{Anc\} of \{Class\}"}, where $\{\textit{Anc}\}$ represents the learnable anchor token. The preposition ``\textit{of}" is used to associate the belonging relationship between the anchor and the class. Through the supervision of the broad category description text, the anchor is encouraged to learn the common internal representation across categories. In the following stage, anchors are frozen and integrated into the normal learnable prompt, then jointly optimized with the position matrix for downstream tasks.

To ensure compatibility with diverse architectures, we propose two depth-adaptive variants (shallow/deep). As a plug-and-play module, AnchorOPT can seamlessly replace existing templates (e.g., CoOp~\cite{zhou2022learning}, ATPrompt~\cite{li2025advancing}), yielding consistent gains across multiple benchmarks. Notably, AnchorOPT achieves superior baseline performance compared to other conventional templates, reaching performance comparable to or better than many methods~\cite{zhou2022conditional,lu2022prompt,yao2023visual,khattak2023maple,xie2025textrefiner}, which incorporate additional regularization or learnable modules. This demonstrates significant underutilized potential in fundamental prompt structures.

In summary, the contributions are listed as follows:
\begin{itemize}
    \item  We introduce AnchorOPT, a dynamic anchor-based prompt learning framework for CLIP that transforms anchors from explicit static tokens into implicitly optimized representations.
    \item A learnable position matrix is introduced to enable task-specific prompt restructuring via differentiable token reordering.
    \item Two depth-compatible variants~(shallow/deep) are introduced to facilitate integration with various prompt learning architectures.
    \item Extensive experiments demonstrate the effectiveness of our AnchorOPT on various datasets.
\end{itemize}

\section{Related Work}
\textbf{Prompt Learning for VLMs.}
Prompt learning~\cite{zhou2022learning,zhou2022conditional,khattak2023maple,lee2023read,khattak2023self,li2024promptkd} has emerged as a parameter-efficient tool~\cite{hu2022lora,gao2024clip} to adapt pre-trained VLMs to downstream tasks. This approach adds a small number of learnable embeddings alongside model input, which are optimized during training while keeping the encoder parameters frozen. After training, models can achieve performance parity with, or even outperform, fully fine-tuned ones~\cite{jia2022visual,zhou2022learning}.
CoOp~\cite{zhou2022learning} is the pioneering method for CLIP, which proposed replacing the hand-crafted hard prompt
with learnable soft prompts for fine-tuning while freezing the entire pre-trained parameters. Subsequent works have adopted this prompt format and designed a variety of learning methods. 
CoCoOp~\cite{zhou2022conditional} aims to enhance the representations of soft tokens by utilizing input images. 
MaPLe~\cite{khattak2023maple} appends the soft tokens to the intermediate hidden representations in both the text and the image encoders. 
RPO~\cite{lee2023read} treats the learnable soft tokens of the two modalities as separate classification tokens and averages them to obtain the final prediction result. 
DePT~\cite{zhang2024dept} decouples the representations of base classes and new classes, introducing additional modules to enhance each. 
PromptKD~\cite{li2024promptkd} leverages a large pre-trained teacher model to provide pseudo labels to supervise the CLIP student with learnable prompts on unlabeled domain data. 

\noindent\textbf{Prompt Learning with Additional Anchors.}
Existing prompt learning methods~\cite{zhou2022learning,zhou2022conditional,khattak2023maple} primarily build upon the cascade structure of multiple soft tokens and class tokens proposed by CoOp~\cite{zhou2022learning}, which serves as input to the encoder during optimization. However, ATPrompt~\cite{li2025advancing} observes that this structure confines soft tokens to learning category-specific representations, thereby disrupting their association with unknown categories. 
Several studies~\cite{zheng2024large,yao2024tcp,tian2024argue,li2025advancing} have introduced additional tokens as anchors to guide soft tokens toward learning more generalizable representations.
These anchor tokens are typically derived from text embeddings. 
For instance, ArGue~\cite{tian2024argue} builds a pool of intra-class attributes by querying an LLM and appends each attribute to the input prompt for joint optimization, aligning text embeddings with attribute groups. 
TCP~\cite{yao2024tcp} proposes embedding tokens encoded by hand-crafted templates as anchors in the text encoder's intermediate layers. While previous methods focused on intermediate layers with complex designs, ATPrompt~\cite{li2025advancing} adopts a fundamental approach by proposing attribute-based prompt learning templates that utilize individual attributes as anchor terms, effectively enhancing model generalization. This refocuses research attention on reconsidering the basic forms of prompt learning.

However, these textual anchors exhibit limitations: their content relies on human-defined attributes, and their position remains fixed,
overlooking the need for dynamic adaptation of input structure across different tasks or learning stages. To address these issues, we propose AnchorOPT, a dynamic anchor-based prompt learning framework. Specifically, AnchorOPT converts explicit, fixed anchor tokens into implicitly learnable components and introduces a learnable position matrix to dynamically adjust token ordering based on training progress. 


\section{Method}


\subsection{Preliminary}
\label{section:preliminary}

\textbf{Vision-Language Models.} 
Existing vision-language models like CLIP~\cite{radford2021learning} and ALIGN~\cite{jia2021scaling} typically have two parallel encoders, one for image and the other for text, which are denoted as $f_{Img}(\cdot)$ and $f_{T}(\cdot)$, respectively. Let $D=\{(x_{i}, y_{i})\}_{i=1}^{I}$ be the training dataset, where $x_{i}$ is the $i$-th input data sample, $y_{i}$ is the corresponding ground truth label, $I$ is the size of dataset and $y_{i}\in \{1, 2, ..., C\}$ for $C$-class classification problem. For each input image $x$, it is first split into $P$ patches, which are then projected to create patch tokens. The image encoder encodes the input patch through the transformer block to produce the visual features $u=f_{I}(x)/ || f_{I}(x) ||_{2} \in \mathbb{R}^{d}$, where $d$ represents the feature dimension.
On the text side, the text label $y$ and the associated name are formulated with the template, e.g., $t_{c}$ = \textit{``A photo of a \{Class $c$\}"}. The text encoder then encodes the input text through the transformer block, generating the textual feature $v_{c}=f_{T}(t_{c})/ ||f_{T}(t_{c})||_{2} \in \mathbb{R}^{d}$. Finally, the probability of $c$-th class for $x_{i}$ is computed as:
\begin{equation}
    q(c|x_{i}) = \frac{\exp(u v_{c}^{\mathsf{T}}/\tau)}{\sum_{j=1}^{C}\exp(u v^{\mathsf{T}}_{j}/\tau)}.
\label{equation:output_prob}
\end{equation}
where $\tau$ denotes the temperature parameter.

\noindent\textbf{Prompt Learning.}
Traditional methods often rely on the generic template as a text prompt for calculation. While this approach works for basic tasks, it lacks specificity for domain- or task-specific applications, failing to yield accurate predictions due to its overgeneralized formulation. Recent works~\cite{zhou2022learning,zhou2022conditional} propose to replace fixed text with implicitly learnable soft tokens, which are optimized for downstream tasks while keeping the encoder parameters frozen. Specifically, $M$ soft tokens 
$[V_{1}], \ldots, [V_{M}]$
are concatenated with fixed class token $[\text{CLS}]$ and input into the text encoder, as shown in Fig.~\ref{fig:compare} (a). Its form can be expressed as follows:
\begin{equation}
    t_{v}=[V_{1}][V_{2}]\ldots[V_{M}][\text{CLS}].
\label{equation:origin_prompt}
\end{equation}
where $M$ represents the soft prompt length. We omit the prefix and suffix tokens for simplicity.

Subsequently, anchor-based methods~\cite{chen2023ovarnet,zheng2024large,tian2024argue,li2025advancing} were proposed to integrate additional tokens into prompts at fixed positions as anchors, guiding soft tokens to learn general and category-invariant representations. Anchor tokens are typically derived from explicit textual patterns, such as class attributes, which are encoded into the token values. Taking ATPrompt~\cite{li2025advancing} as an example, its prompt structure is formulated as follows:
\begin{equation}
    t_{attr}=[V_{a_{1}}]\ldots[V_{a_{m}}][\text{Anc}^{e}][V_{1}]\ldots[V_{M}][\text{CLS}].
\label{equation:attribute_prompt}
\end{equation}
where $[\text{Anc}^{e}]$ represents the explicitly encoded anchor token obtained and $[V]$ represent the learnable soft tokens.

\begin{figure*}[t]
    \centering
    \includegraphics[width=0.99\linewidth]{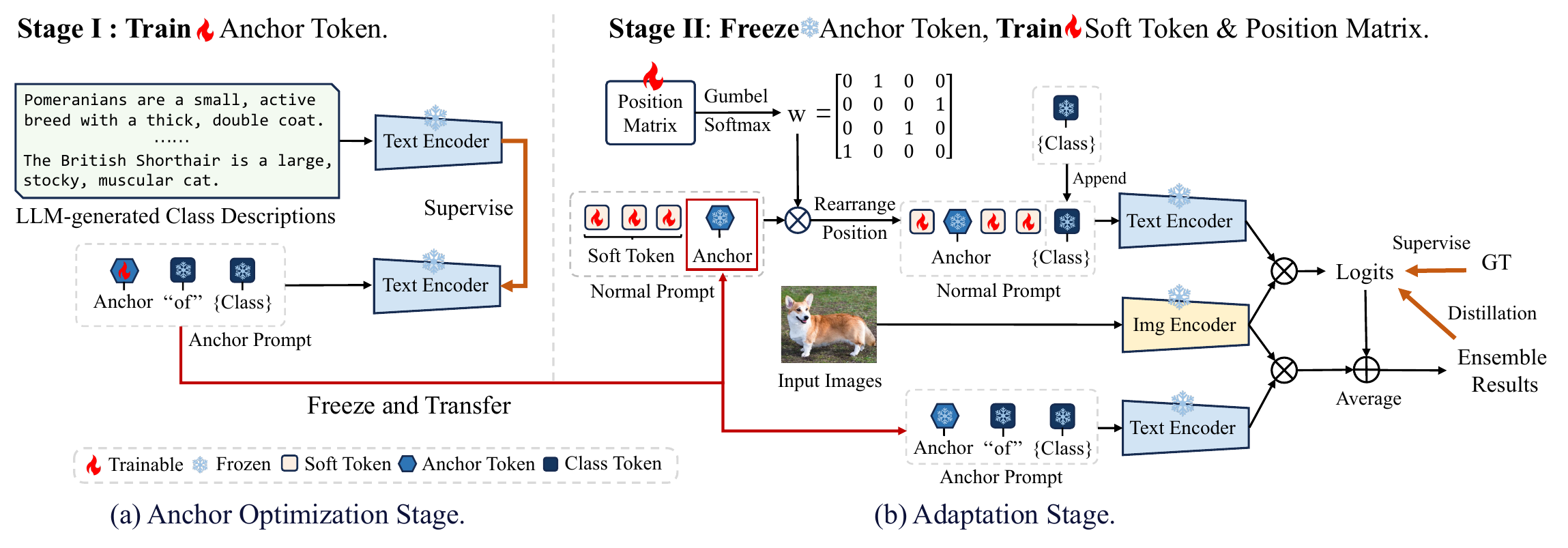}
    \vspace{-8pt}
    \caption{
    The dynamic anchor token training process comprises two stages: (a) Anchor Optimization: Anchor tokens are initialized as learnable parameters and optimized using LLM-generated category descriptions. The resulting anchors are frozen for the subsequent stage. (b) Adaptation: Soft prompts and the position matrix are jointly optimized for downstream tasks, with knowledge distillation from ensemble results providing auxiliary supervision.
    }
    \label{fig:framework}
    \vspace{-5pt}
\end{figure*}

\subsection{Prompt Learning with Dynamic Anchor Token}
\label{section:pipeline}

Existing anchor-based prompt learning methods like ATPrompt rely on explicitly encoding text (e.g., ``shape", ``color") as anchor values and fixing their positions within input prompts. This approach imposes human-defined explicit text as token values, which may not best meet the actual needs of downstream tasks. Furthermore, the static nature of these anchors—both in value and position—prevents adaptive refinement during training, limiting their ability to accommodate task-specific needs.


To address these limitations, we propose AnchorOPT, a dynamic anchor-based method for prompt learning, as illustrated in Fig.~\ref{fig:compare}(d). The dynamic nature of anchor tokens manifests in two key aspects: (1)~\textit{Value Dynamics}: Anchor token values are learned dynamically from data rather than being manually predefined and kept fixed throughout.
(2)~\textit{Position Dynamics}: The positions of both soft and anchor tokens are adaptively adjusted through a learnable position matrix, instead of remaining fixed during training. 
To effectively train and utilize these anchors, we propose a two-stage framework comprising anchor optimization and adaptation. Each stage incorporates the corresponding dynamic aspect. Detailed explanations of both stages follow.

 

\subsubsection{Stage I: Anchor Optimization}

\textbf{Dynamic Values.}
Prior work~\cite{li2025advancing} demonstrates that anchors encoding explicit cross-category attributes (e.g., ``shape", ``color") bridge generalization gaps between known and unknown classes. Extending this principle, we propose a specialized anchor prompt template: $t_{anc}$= \textit{``\{Anc\} of \{Class\}."}, where learnable anchor tokens implicitly aggregate cross-category semantics. The preposition \textit{``of"} models associative relationships between anchors and classes. Specifically, the anchor prompt can be formalized as:
\begin{equation}
    t_{anc}=[\text{Anc}^{i}_{1}]...[\text{Anc}^{i}_{N}][\text{of}][\text{CLS}],
\end{equation}
where $[\text{Anc}^{i}]$ represents the implict learnable anchor token, $N$ is the anchor token count, $[\text{of}]$ is the fixed preposition embedding, and $[\text{CLS}]$ represents the class token. This transforms anchors from explicitly defined static tokens into implicitly optimized representations.

\begin{figure*}
    \centering
    \includegraphics[width=0.85\linewidth]{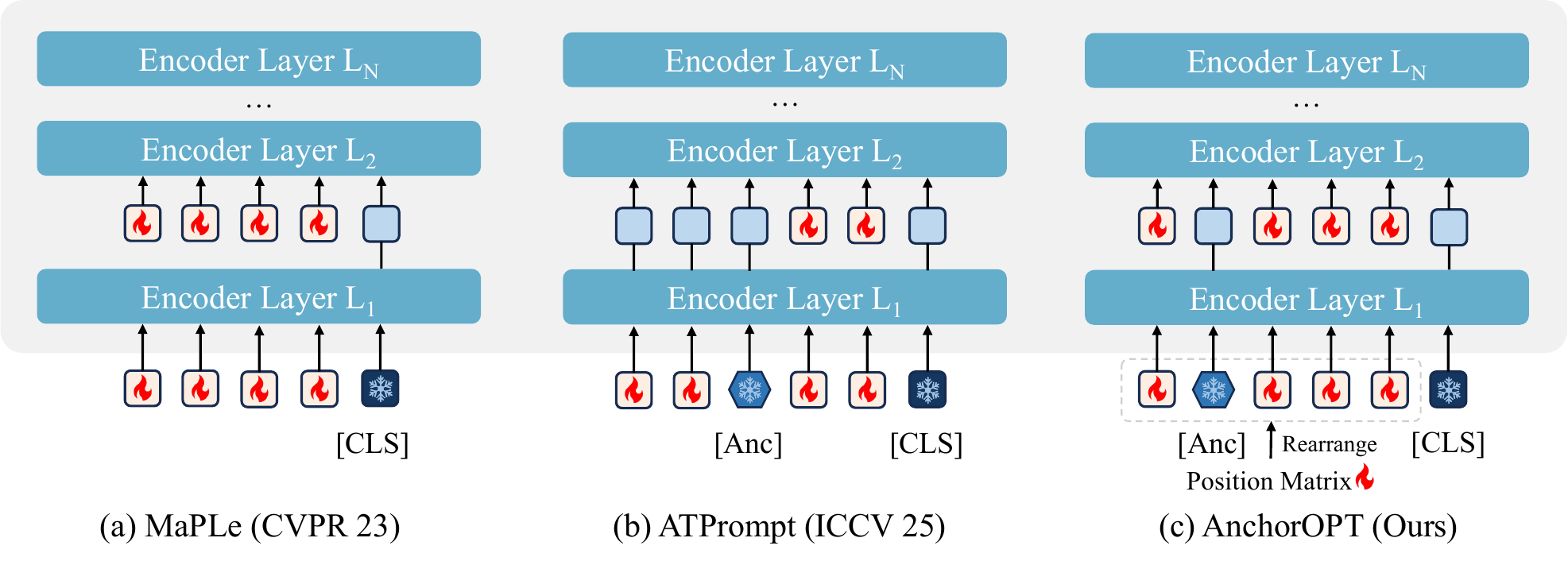}
    \vspace{-10pt}
    \caption{Computational process in deep prompt learning variants: (a) MaPLe drops and reintroduces all soft tokens after each Transformer block. (b) ATPrompt retains all attribute-related hard/soft tokens while discarding class-related soft tokens. (c) AnchorOPT dynamically reorders tokens via the position matrix, retaining only anchor tokens and discarding all soft tokens during processing.}
    \vspace{-10pt}
    \label{figure:deep_version}
\end{figure*}

\noindent\textbf{Training.}
To enable anchor tokens to capture universal cross-category representations, we align the anchor prompt $t_{anc}$ with LLM-generated category descriptions $t_{d}$~(Fig.~\ref{fig:framework}(a)) using Mean Squared Error~(MSE) loss. The objective function can be formulated as:
\begin{equation}
    L_{\theta_{a}} = \text{MSE}(f_{T}(t_{anc}; \theta_{a}), f_{T}(t_{d})).
\label{equation:anchor_mse}
\end{equation}
where $\theta_{a}$ denotes the anchor token parameters and $f_{T}(\cdot)$ is the text encoder.
Following anchor optimization, the $N$-token anchor sequence is concatenated with an $M$-token soft prompt to form the integrated prompt for downstream adaptation.



\subsubsection{Stage II: Adaptation}
\textbf{Dynamic Positions.} Inspired by how humans adapt sentence structures to meet communicative needs across different contexts, we argue that the fixed structures used in prior work may constrain model adaptability to downstream tasks. To enable adaptive sequence adjustment of the normal prompt, we introduce a learnable two-dimensional position matrix $W$ of size $(M+N, M+N)$ which dynamically determines the relative positions between soft tokens and anchor tokens, as shown in Fig.~\ref{fig:framework}(b). Here, $W_{i}^{j}$ indicates the probability that the $j$-th token in the original sequence is assigned to the $i$-th position in the rearranged sequence. However, since the assignment operation is inherently discrete, the matrix $W$ is non-differentiable.

To enable differentiable learning, we adopt the Gumbel-Softmax relaxation~\cite{jang2016categorical}. 
Specifically, for each target position $i$ in the output sequence, we define a probability vector $p^{i}=[p_{1}^{i}, ..., p_{N+M}^{i}]$, where $p_{j}^{i}$ denotes the probability that token $j$ is selected for position $i$. During the forward pass, the discrete assignment is obtained by applying the argmax operation with Gumbel noise:
\begin{equation} 
\mathbf{w}^{i} = \text{one\_hot}\{\arg\max_{j} (\log p_{j}^{i} + \epsilon_{j}) \},
\end{equation} 
where $\epsilon_{j} \sim \text{Gumbel}~(0, 1)$ are independent samples from the standard Gumbel distribution, and $\text{one\_hot}(\cdot)$ converts the index into a one-hot vector.

During training, to allow gradient backpropagation, we approximate the discrete assignment using a continuous softmax:
\begin{equation} 
\mathbf{w}^{i}_{j} = \frac{\exp\left( (\log p^i_j + \epsilon_j) / \tau \right)}{\sum_{k=1}^{N+M} \exp\left( (\log p^i_k + \epsilon_k) / \tau \right)}. 
\end{equation}
where the temperature parameter $\tau$ controls the softmax sharpness. 
The resulting softened weight matrix $\hat{W}$ is used for gradient computation, while the hard assignment version is applied during inference.
Subsequently, the frozen anchor tokens are concatenated with soft tokens to form the composite prompt $t_{v}$, which is then restructured via the position matrix before appending the class token $[\text{CLS}]$ for encoder input.
This process can be formally expressed as:
\begin{equation}
t_v = [V_{1}]...[V_{M}][\text{Anc}^{i}_{1}]...[\text{Anc}^{i}_{N}],
\end{equation}
\begin{equation}
t_{norm} = \operatorname{concat}( \hat{W} \odot t_{v}, [\text{CLS}]).
\label{equation:positioned_prompt}
\end{equation}
The resulting $t_{norm}$ is then passed to the text encoder as a structured prompt for downstream processing.

\noindent\textbf{Training.} 
In the second stage, two prompts are used: the trainable normal prompt $t_{norm}$ and the pre-trained anchor prompt $t_{anc}$ from the first stage. For the normal prompt, we optimize the learnable position matrix and soft tokens using the conventional cross-entropy loss. Let $\theta_{s}$ and $\theta_{pos}$ denote the parameters of the soft tokens and position matrix, respectively. The loss function $L_{\theta_{s}, \theta_{pos}}$ is formulated as:
\begin{equation}
    L^{1}_{\theta_{s}, \theta_{pos}} = \text{CE}(q_{norm}, y),
\label{equation:ce_downstream}
\end{equation}
where $q_{norm}=f(x, t_{norm}; \theta_{s}, \theta_{p})$ denotes the output of the CLIP model given image $x$ and prompt $t_{norm}$.

For the anchor prompt, we first compute its prediction $q_{anc}$ for the input image. This prediction is then combined with the normal prompt's output $q_{norm}$ through averaging to form an ensemble prediction $q_{ens}$. We then distill the knowledge~\cite{hinton2015distilling,zhu2018knowledge,li2023curriculum} from this ensemble output to the learnable normal prompt using the KL divergence loss:
\begin{equation}
    L^{2}_{\theta_{s}, \theta_{pos}} = \text{KL}(q_{norm}, q_{ens}),
\end{equation}
where $\text{KL}(\cdot, \cdot)$ denotes the Kullback–Leibler divergence.

The overall training objective for the second stage is:
\begin{equation}
    L_{total} = \lambda_{1} \cdot L^{1}_{\theta_{s}, \theta_{pos}} + \lambda_{2} \cdot L^{2}_{\theta_{s}, \theta_{pos}}.
\end{equation}
where $\lambda_{1}$ and $\lambda_{2}$ are hyperparameters that control the loss balance, typically set to a 1:10 ratio.

\noindent\textbf{Inference.} During inference, we use predictions from well-trained normal prompts for known base classes, while employing ensemble results for unknown novel classes.



\begin{table*}[t]
    \begin{subtable}{\textwidth}
    \centering
    \resizebox{0.81\linewidth}{!}
    {
    \centering
    \begin{tabular}{cccc|ccc|ccc|ccc}
    \toprule 
    \multirow{2}[3]{*}{Method} & \multicolumn{3}{c}{Average} & \multicolumn{3}{c}{ImageNet} & \multicolumn{3}{c}{ Caltech101} & \multicolumn{3}{c}{OxfordPets} \\
    \cmidrule(lr){2-4}\cmidrule(lr){5-7}\cmidrule(lr){8-10}\cmidrule(lr){11-13}
    & Base & Novel & HM & Base & Novel & HM & Base & Novel & HM & Base & Novel & HM \\
    \midrule
    PromptSRC~\scriptsize{\textcolor{gray}{(ICCV 23)}} & 84.26 & 76.10 & 79.97 & 77.60 & 70.73 & 74.01 & 98.10 & 94.03 & 96.02 & 95.33 & 97.30 & 96.30 \\
    CoPrompt~\scriptsize{\textcolor{gray}{(ICLR 24)}} & 84.00 & 77.23 & 80.48 & 77.67 & 71.27 & 74.33 & 98.27 & 94.90 & 96.55 & 95.67 & 98.10 & 96.87 \\
    TCP~\scriptsize{\textcolor{gray}{(CVPR 24)}} & 84.13 & 75.36 & 79.51 & 77.27 & 69.87 & 73.38 & 98.23 & 94.67 & 96.42 & 94.67 & 97.20 & 95.92 \\
    TextRefiner~\scriptsize{\textcolor{gray}{(AAAI 25)}} & 79.74 & 74.32 & 76.94 & 76.84 & 70.54 & 73.56 & 98.13 & 94.43 & 96.24 & 95.27 & 97.65 & 96.45 \\
    DPC~\scriptsize{\textcolor{gray}{(CVPR 25)}} & 85.15 & 68.84 & 76.13 & 77.72 & 68.85 & 73.02 & 98.58 & 94.65 & 96.58 & 95.80 & 97.60 & 96.69 \\
    \midrule
    CoOp~\scriptsize{\textcolor{gray}{(IJCV 22)}} & 82.69 & 63.22 & 71.66 & 76.47 & 67.88 & 71.92 & 98.00 & 89.81 & 93.73 & 93.67 & 95.29 & 94.47 \\
    + ATPrompt~\scriptsize{\textcolor{gray}{(ICCV 25)}} & 82.68 & 68.04 & 74.65~\scriptsize{\textcolor{gray}{(+2.99)}}  & 76.27 & 70.60 & 73.33 & 97.95 & 93.63 & 95.74 & 94.77 & 96.59 & 95.67 \\ 
    + AnchorOPT~(Ours) & 81.24 & 76.27 & \textbf{78.68}~\scriptsize{\incre{(+7.02)}} & 76.30 & 71.56 & \textbf{73.85} & 98.26 & 95.34 & \textbf{96.78} & 95.32 & 98.04 & \textbf{96.66} \\
    \midrule
    CoCoOp~\scriptsize{\textcolor{gray}{(CVPR 22)}} & 80.47 & 71.69 & 75.83 & 75.98 & 70.43 & 73.10 & 97.96 & 93.81 & 95.84 & 95.20 & 97.69 & 96.43 \\
    + ATPrompt~\scriptsize{\textcolor{gray}{(ICCV 25)}} & 81.69 & 74.54 & 77.95~\scriptsize{\textcolor{gray}{(+2.21)}} & 76.43 & 70.50 & 73.35 & 97.96 & 95.27 & 96.60 & 95.46 & 97.89 & 96.66 \\
    + AnchorOPT~(Ours) & 81.87 & 77.06 & \textbf{79.39}~\scriptsize{\incre{(+3.56)}} & 76.32 & 71.78 & \textbf{73.98} & 98.17 & 95.56 & \textbf{96.85} & 95.87 & 98.06 & \textbf{96.95} \\
    \midrule
    MaPLe~\scriptsize{\textcolor{gray}{(CVPR 23)}} & 82.28 & 75.14 & 78.55 & 76.66 & 70.54 & 73.47 & 97.74 & 94.36 & 96.02 & 95.43 & 97.76 & 96.58 \\
    + ATPrompt~\scriptsize{\textcolor{gray}{(ICCV 25)}} & 82.98 & 75.76 & 79.21~\scriptsize{\textcolor{gray}{(+0.66)}} & 76.94 & 70.72 & 73.70 & 98.32 & 95.09 & 96.68 & 95.62 & 97.63 & 96.61 \\
    + AnchorOPT~(Ours) & 83.62 & 77.36 & \textbf{80.37}~\scriptsize{\incre{(+1.82)}} & 76.98 & 71.88 & \textbf{74.34} & 97.87 & 96.03 & \textbf{96.94} & 96.38 & 98.23 & \textbf{97.30} \\
    \midrule   
    DePT~\scriptsize{\textcolor{gray}{(CVPR 24)}} & 83.66 & 71.82 & 77.29 & 77.13 & 70.10 & 73.45 & 98.33 & 94.33 & 96.29 & 94.70 & 97.63 & 96.14 \\
    + ATPrompt~\scriptsize{\textcolor{gray}{(ICCV 25)}} & 83.80 & 73.75 & 78.45~\scriptsize{\textcolor{gray}{(+1.16)}} & 77.32 & 70.65 & 73.83 & 98.48 & 94.60 & 96.50 & 94.65 & 97.99 & 96.29 \\
    + AnchorOPT~(Ours) & 84.27 & 76.90 & \textbf{80.42}~\scriptsize{\incre{(+3.13)}} & 77.56 & 71.67 & \textbf{74.50} & 98.20 & 95.78 & \textbf{96.97} & 95.39 & 98.23 & \textbf{96.79} \\
    \bottomrule
    \end{tabular}
    }
    \end{subtable}

    \vspace{5pt}
    \begin{subtable}{\textwidth}
    \centering
    \resizebox{0.76\linewidth}{!}
    {
    \begin{tabular}{cccc|ccc|ccc|ccc}
    \toprule \multirow{2}[3]{*}{ Method } & \multicolumn{3}{c}{StanfordCars} & \multicolumn{3}{c}{ Flowers102} & \multicolumn{3}{c}{ Food101} & \multicolumn{3}{c}{ FGVCAircraft} \\
    \cmidrule(lr){2-4}\cmidrule(lr){5-7}\cmidrule(lr){8-10}\cmidrule(lr){11-13} 
    & Base & Novel & HM & Base & Novel & HM & Base & Novel & HM & Base & Novel & HM \\
    \midrule
    PromptSRC~\scriptsize{\textcolor{gray}{(ICCV 23)}} & 78.27 & 74.97 & 76.58 & 98.07 & 76.50 & 85.95 & 90.67 & 91.53 & 91.10 & 42.73 & 37.87 & 40.15 \\
    CoPrompt~\scriptsize{\textcolor{gray}{(ICLR 24)}} & 76.97 & 74.40 & 75.66 & 97.27 & 76.60 & 85.71 & 90.73 & 92.07 & 91.40 & 40.20 & 39.33 & 39.76 \\
    TCP~\scriptsize{\textcolor{gray}{(CVPR 24)}} & 80.80 & 74.13 & 77.32 & 97.73 & 75.57 & 85.23 & 90.57 & 91.37 & 90.97 & 41.97 & 34.43 & 37.83 \\
    TextRefiner~\scriptsize{\textcolor{gray}{(AAAI 25)}} & 71.40 & 70.90 & 71.15 & 95.92 & 74.33 & 83.76 & 90.88 & 91.43 & 91.15 & 35.35 & 35.87 & 35.61 \\
    DPC~\scriptsize{\textcolor{gray}{(CVPR 25)}} & 81.13 & 70.14 & 75.24 & 98.86 & 68.37 & 80.84 & 91.15 & 91.47 & 91.31 & 45.56 & 24.24 & 31.64 \\
    \midrule
    CoOp~\scriptsize{\textcolor{gray}{(IJCV 22)}} & 78.12 & 60.40 & 68.13 & 97.60 & 59.67 & 74.06 & 88.33 & 82.26 & 85.19 & 40.44 & 22.30 & 28.75 \\
    + ATPrompt~\scriptsize{\textcolor{gray}{(ICCV 25)}} & 77.43 & 66.55 & 71.58 & 97.44 & 67.52 & 79.77 & 88.74 & 87.44 & 88.09 & 40.38 & 27.22 & 32.52 \\
    + AnchorOPT (Ours) & 76.73 & 74.43 & \textbf{75.67} & 97.22 & 75.41 & \textbf{84.94} & 90.40 & 91.95 & \textbf{91.17} & 38.48 & 35.47 & \textbf{36.91} \\
    \midrule
    CoCoOp~\scriptsize{\textcolor{gray}{(CVPR 22)}} & 70.49 & 73.59 & 72.01 & 94.87 & 71.75 & 81.71 & 90.70 & 91.29 & 90.99 & 33.41 & 23.71 & 27.74 \\
    + ATPrompt~\scriptsize{\textcolor{gray}{(ICCV 25)}} & 74.50 & 73.47 & 73.98 & 96.52 & 73.59 & 83.51 & 90.59 & 91.74 & 91.16 & 37.30 & 33.15 & 35.10 \\
    + AnchorOPT~(Ours) & 75.59 & 75.23 & \textbf{75.41} & 96.17 & 77.35 & \textbf{85.74} & 90.58 & 91.98 & \textbf{91.27} & 36.90 & 36.35 & \textbf{36.62} \\
    \midrule
    MaPLe~\scriptsize{\textcolor{gray}{(CVPR 23)}} & 72.94 & 74.00 & 73.47 & 95.92 & 72.46 & 82.56 & 90.71 & 92.05 & 91.38 & 37.44 & 35.61 & 36.50 \\
    + ATPrompt~\scriptsize{\textcolor{gray}{(ICCV 25)}} & 75.39 & 73.84 & 74.61 & 97.82 & 75.07 & 84.95 & 90.65 & 92.00 & 91.32 & 37.61 & 36.15 & 36.87 \\
    + AnchorOPT~(Ours) & 77.30 & 74.49 & \textbf{75.87} & 97.79 & 77.35 & \textbf{86.38} & 90.58 & 92.16 & \textbf{91.36} & 38.72 & 38.19 & \textbf{38.45} \\
    \midrule
    DePT~\scriptsize{\textcolor{gray}{(CVPR 24)}} & 79.67 & 72.40 & 75.86 & 98.20 & 72.00 & 83.08 & 90.43 & 91.33 & 90.88 & 42.53 & 22.53 & 29.46 \\
    + ATPrompt~\scriptsize{\textcolor{gray}{(ICCV 25)}} & 79.29 & 73.47 & 76.27 & 98.20 & 73.69 & 84.20 & 90.42 & 91.69 & 91.05 & 43.19 & 33.23 & 37.56 \\
    + AnchorOPT~(Ours) & 80.98 & 75.63 & \textbf{78.21} & 98.10 & 78.30 & \textbf{87.09} & 90.53 & 91.93 & \textbf{91.22} & 44.34 & 36.09 & \textbf{39.79} \\
    \bottomrule
    \end{tabular}
    }
    \end{subtable}

    \vspace{5pt}
    \begin{subtable}{\textwidth}
    \centering
    \resizebox{0.76\linewidth}{!}
    {
    \begin{tabular}{cccc|ccc|ccc|ccc}
    \toprule 
    \multirow{2}[3]{*}{Method} & \multicolumn{3}{c}{SUN397} & \multicolumn{3}{c}{DTD} & \multicolumn{3}{c}{EuroSAT} & \multicolumn{3}{c}{UCF101} \\
    \cmidrule(lr){2-4}\cmidrule(lr){5-7}\cmidrule(lr){8-10}\cmidrule(lr){11-13} 
    & Base & Novel & HM & Base & Novel & HM & Base & Novel & HM & Base & Novel & HM \\
    \midrule
    PromptSRC~\scriptsize{\textcolor{gray}{(ICCV 23)}} & 82.67 & 78.47 & 80.52 & 83.37 & 62.97 & 71.75 & 92.90 & 73.90 & 82.32 & 87.10 & 78.80 & 82.74 \\
    CoPrompt~\scriptsize{\textcolor{gray}{(ICLR 24)}} & 82.63 & 80.03 & 81.30 & 83.13 & 64.73 & 72.79 & 94.60 & 78.57 & 85.84 & 86.90 & 79.57 & 83.07 \\
    TCP~\scriptsize{\textcolor{gray}{(CVPR 24)}} & 82.63 & 78.20 & 80.35 & 82.77 & 58.07 & 68.25 & 91.63 & 74.73 & 82.32 & 87.13 & 80.77 & 83.83 \\
    TextRefiner~\scriptsize{\textcolor{gray}{(AAAI 25)}} & 80.96 & 76.49 & 78.66 & 75.35 & 58.09 & 65.60 & 74.57 & 72.82 & 73.68 & 82.52 & 75.01 & 78.59\\
    DPC~\scriptsize{\textcolor{gray}{(CVPR 25)}} & 82.81 & 74.10 & 78.21 & 84.61 & 49.88 & 62.76 & 93.40 & 51.62 & 66.49 & 87.02 & 66.31 & 75.27 \\
    \midrule
    CoOp~\scriptsize{\textcolor{gray}{(IJCV 22)}} & 80.60 & 65.89 & 72.51 & 79.44 & 41.18 & 54.24 & 92.19 & 54.74 & 68.69 & 84.69 & 56.05 & 67.46 \\
    + ATPrompt~\scriptsize{\textcolor{gray}{(ICCV 25)}} & 80.84 & 68.64 & 74.24 & 80.83 & 45.49 & 58.22 & 90.34 & 59.79 & 71.96 & 84.49 & 64.96 & 73.45 \\
    + AnchorOPT (Ours) & 79.67 & 78.73 & \textbf{79.20} & 74.04 & 61.72 & \textbf{67.32} & 83.75 & 77.14 & \textbf{80.31} & 83.45 & 79.18 & \textbf{81.26} \\
    \midrule
    + CoCoOp~\scriptsize{\textcolor{gray}{(CVPR 22)}} & 79.74 & 76.86 & 78.27 & 77.01 & 56.00 & 64.85 & 87.49 & 60.04 & 71.21 & 82.33 & 73.45 & 77.64 \\
    ATPrompt~\scriptsize{\textcolor{gray}{(ICCV 25)}} & 80.50 & 76.86 & 78.64 & 78.63 & 56.89 & 66.02 & 87.95 & 74.15 & 80.46 & 82.74 & 76.40 & 79.44 \\
    + AnchorOPT~(Ours) & 79.74 & 79.25 & \textbf{79.49} & 77.43 & 63.29 & \textbf{69.65} & 89.54 & 79.51 & \textbf{84.23} & 83.27 & 79.28 & \textbf{81.23} \\
    \midrule
    MaPLe~\scriptsize{\textcolor{gray}{(CVPR 23)}} & 80.82 & 78.70 & 79.75 & 80.36 & 59.18 & 68.16 & 94.07 & 73.23 & 82.35 & 83.00 & 78.66 & 80.77 \\
    + ATPrompt~\scriptsize{\textcolor{gray}{(ICCV 25)}} & 80.98 & 78.15 & 79.54 & 80.50 & 58.28 & 67.61 & 94.84 & 77.59 & 85.35 & 84.08 & 78.88 & 81.40 \\
    + AnchorOPT~(Ours) & 81.99 & 79.57 & \textbf{80.76} & 81.94 & 60.94 & \textbf{69.90} & 95.92 & 81.63 & \textbf{88.20} & 84.35 & 80.53 & \textbf{82.40} \\
    \midrule
    DePT~\scriptsize{\textcolor{gray}{(CVPR 24)}} & 82.37 & 75.07 & 78.55 & 83.20 & 56.13 & 67.04 & 88.27 & 66.27 & 75.70 & 85.43 & 72.17 & 78.24 \\
    + ATPrompt~\scriptsize{\textcolor{gray}{(ICCV 25)}} & 82.42 & 76.48 & 79.34 & 82.64 & 56.77 & 67.30 & 89.60 & 69.50 & 78.28 & 85.60 & 73.15 & 78.89 \\
    + AnchorOPT~(Ours) & 82.41 & 79.08 & \textbf{80.71} & 81.87 & 63.16 & \textbf{71.31} & 91.12 & 77.80 & \textbf{83.93} & 86.42 & 78.27 & \textbf{82.14} \\
    \bottomrule
    \end{tabular}
    }
    \end{subtable}
    \vspace{-5pt}
    \caption{Base-to-novel generalization experiments of various baselines with and without AnchorOPT on 11 datasets. As a foundational template-based method, AnchorOPT demonstrates superior performance compared to ATPrompt. Notably, equipped with our AnchorOPT, the current simple method exhibits robust capabilities, even surpassing some complex, advanced methods that introduce additional regularization techniques or learning modules.
    }
    \label{table:base_to_novel}
    \vspace{-5pt}
\end{table*}

\subsubsection{Deep Version of AnchorOPT}

Beyond incorporating soft tokens solely in the input layer~(shallow variant), we extend this approach by introducing learnable soft tokens in deeper transformer layers, as depicted in Fig.~\ref{figure:deep_version}(c). Initially, a position matrix dynamically reorders soft and anchor tokens, which are then concatenated with the class token for network input. Unlike existing methods~\cite{khattak2023maple,khattak2023self,li2024promptkd} that discard all soft tokens after each transformer block’s forward pass and reintroduce them before subsequent blocks, our method preserves anchor tokens throughout deep-layer processing to maintain critical semantic guidance. This retention enables anchor tokens to influence deeper representations while soft tokens are selectively discarded and reintroduced between layers.


\begin{table*}[ht]
    \centering
    \resizebox{0.85\linewidth}{!}
    {
        \begin{tabular}{ccccccccccccccc}
        \toprule
        \multirow{3}[3]{*}{Method} & Source & \multicolumn{10}{c}{Target Dataset} & \multirow{3}[3]{*}{Average} \\
        \cmidrule(lr){2-2}\cmidrule(lr){3-12}
        ~ & Image & Caltech & Oxford & Stanford & Flowers & Food & FGVC     & SUN & \multirow{2}*{DTD} & Euro & UCF &  \\
        ~ & Net   & 101     & Pets   & Cars     & 102     & 101  & Aircraft & 397 & ~     & SAT  & 101 & ~ \\
        \midrule
        CoCoOp & 71.02 & 94.43 & 90.14 & 65.32 & 71.88 & 86.06 & 22.94 & 67.36 & 45.73 & 45.37 & 68.21 & 65.74 \\
        MaPLe & 70.72 & 93.53 & 90.49 & 65.57 & 72.23 & 86.20 & 24.74 & 67.01 & 46.49 & 48.06 & 68.69 & 66.30 \\
        TCP & 71.40 & 93.97 & 91.25 & 64.69 & 71.21 & 86.69 & 23.45 & 67.15 & 44.35 & 51.45 & 68.73 & 66.29 \\
        CoPrompt & 70.80 & 94.50 & 90.73 & 65.67 & 72.30 & 86.43 & 24.00 & 67.57 & 47.07 & 51.90 & 69.73 & 67.00 \\
        \midrule
        CoOp & 71.51 & 93.70 & 89.14 & 64.51 & 68.71 & 85.30 & 18.47 & 64.15 & 41.92 & 46.39 & 66.55 & 63.88 \\
        + ATPrompt & 71.67 & 93.96 & 90.65 & 65.01 & 70.40 & 85.86 & 20.97 & 65.77 & 43.44 & 46.59 & 69.92 & 65.26~\scriptsize{\textcolor{gray}{(+1.38)}} \\
        + AnchorOPT & 71.33 & 94.65 & 90.08 & 65.90 & 72.39 & 86.51 & 23.13 & 67.99 & 46.93 & 52.10 & 69.60 & \textbf{66.93}~\scriptsize{\incre{(+3.05)}} \\
        
        \bottomrule
        \end{tabular}
    }
    \vspace{-5pt}
    \caption{Cross-dataset generalization experiments on ten datasets. AnchorOPT achieves average performance improvement on target datasets. It's worth noting that simply combining AnchorOPT with CoOp achieves performance comparable to more complex methods like CoPrompt, indicating significant untapped potential in designing basic prompting learning templates.}
    \label{table:cross_dataset}
    \vspace{-5pt}
\end{table*}

\begin{figure*}[t!]
    \centering
    \includegraphics[width=0.92\linewidth]{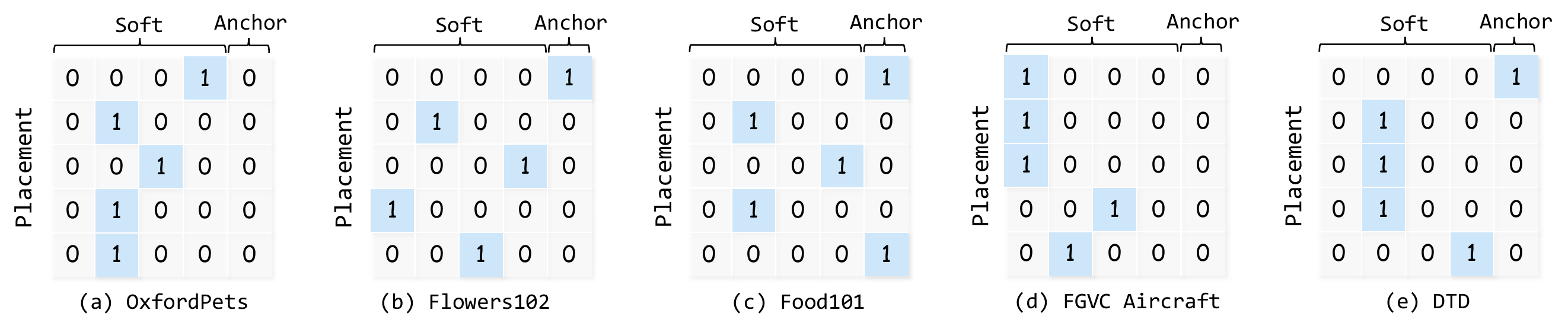}
    \vspace{-7pt}
    \caption{Position matrix visualizations across five datasets. For the Oxford Pets dataset, a value of 1 at row 1, column 4 indicates that token 4 in the original sequence is mapped to position 1 in the transformed sequence. Each dataset exhibits distinct convergence patterns. Note that all-zero values in the final column of the position matrix do not imply anchor tokens are redundant; though omitted from the visualization, these tokens contribute to intermediate computation stages during training.
    }
    \label{fig:visualization}
    \vspace{-10pt}
\end{figure*}

\subsubsection{One-Stage Training Extension}


While our primary framework employs a two-stage training paradigm that decouples anchor tokens from task-specific soft tokens, we propose a computationally efficient one-stage co-training paradigm. This unified approach integrates both training phases into alternating optimization steps within a single framework: (1) updating anchor tokens, and (2) freezing anchor parameters while optimizing soft tokens and the position matrix. As unstable intermediate ensemble results cannot yield reliable supervision signals, we eliminate the distillation objective during one-stage training. More training details are presented in the Appendix.

\section{Experiments}
\subsection{Settings}

\textbf{Baseline Methods.} 
We select four influential prompt learning methods as our baseline and backbone models for our plug-and-play technique, including CoOp~\cite{zhou2022learning}, CoCoOp~\cite{zhou2022conditional}, MaPLe~\cite{khattak2023maple}, and DePT~\cite{zhang2024dept}.
In addition, we also list other methods for comparison, including PromptSRC~\cite{khattak2023self}, CoPrompt~\cite{roy2023consistency}, TCP~\cite{yao2024tcp}, TextRefiner~\cite{xie2025textrefiner}, and DPC~\cite{li2025dpc}.

\noindent\textbf{Implementation Details.}
All experiments were conducted on a single NVIDIA H20 GPU. 
We use the ViT-B/16 CLIP as our default model.
The length of soft tokens is usually set to 4 or 6. The length of the anchor token is set to 1. We report base and novel class accuracy and harmonic mean averaged over 3 runs. 
Complete details are attached in the Appendix.

\subsection{Base-to-Novel Generalization}
\textbf{Results.} In Table~\ref{table:base_to_novel}, we report the quantitative results for base-to-novel tasks on 11 diverse datasets. 
It shows that, combined with our AnchorOPT method, multiple baseline methods have achieved consistent improvements, ranging from 1.82\% to 7.02\%. 
As a fundamental template-based method, AnchorOPT exhibits superior performance compared to ATPrompt.

\noindent\textbf{Position Matrix Visualization.} 
The position matrix learned by our method is visualized in Fig.~\ref{fig:visualization}. It reveals two interesting observations: (1) the matrix sometimes repeatedly selects identical tokens during sequence construction, and (2) tokens exhibit varying importance, with certain tokens discarded in the final stage.


\noindent\textbf{Extension to One-stage Training.} Table~\ref{table:one-stage} compares the average performance of one-stage and two-stage training paradigms using the CoOp method across 11 datasets. The results indicate that while the one-stage approach entails a less complex computational process, it yields lower performance than the two-stage paradigm.

\begin{table}[ht]
    \centering
    \resizebox{0.55\linewidth}{!}
    {
        \begin{tabular}{ccccc}
        \toprule
        Paradigm  & Base  & Novel & HM    \\
        \midrule
        One-stage & 80.36 & 75.89 & 78.06 \\
        Two-stage & 81.24 & 76.27 & \textbf{78.68} \\
        \bottomrule
        \end{tabular}
    }
    \vspace{-5pt}
    \caption{Comparison of training paradigm on 11 datasets. Two-stage approaches yield better results.}
    \label{table:one-stage}
    \vspace{-8pt}
\end{table}

\subsection{Cross-dataset Experiment}
Table~\ref{table:cross_dataset} shows the cross-dataset experimental results on various datasets. Compared to ATPrompt, our method demonstrates stronger cross-dataset generalization ability. 

\subsection{Domain Generalization}

Table~\ref{table:domain_dataset} shows the domain generalization results on four datasets. CoOp+AnchorOPT achieves stronger domain generalization performance than CoPrompt.

\begin{table}[!h]
    \centering
    \resizebox{0.99\linewidth}{!}
    {
        \begin{tabular}{ccccccc}
        \toprule
        \multirow{2}[2]{*}{Method} & Source & \multicolumn{4}{c}{Target Dataset} & \multirow{2}[2]{*}{Average} \\
        \cmidrule(lr){2-2}\cmidrule(lr){3-6}
         ~  & IN & -V2 & -S & -A & -R & ~ \\
         \midrule
        CoCoOp & 71.02 & 64.07 & 48.75 & 50.63 & 76.18 & 59.91 \\
        MaPLe  & 70.72 & 64.07 & 49.15 & 50.90 & 76.98 & 60.27 \\
        TCP & 71.20 & 64.6 & 49.50 & 51.2 & 76.73 & 60.51 \\
        CoPrompt & 70.80 & 64.25 & 49.43 & 50.50 & 77.51 & 60.42 \\
        TextRefiner & 72.06 & 65.02 & 48.58 & 49.77 & 76.30 & 59.92 \\
        \midrule
        CoOp & 71.51 & 64.20 & 47.99 & 49.71 & 75.21 & 59.28 \\
        + ATPrompt & 71.67 & 64.43 & 49.13  & 50.91 & 76.24 & 60.18~\scriptsize{\textcolor{gray}{(+0.90)}} \\
        + AnchorOPT & 71.33 & 64.37 & 49.81 & 51.64 & 77.32 & \textbf{60.79}~\scriptsize{\incre{(+1.51)}} \\
        
        \bottomrule
        \end{tabular}
    }
    \caption{Domain generalization experiments on four datasets. The integration of AnchorOPT resulted in better performance.
    }
    \label{table:domain_dataset}
    \vspace{-8pt}
\end{table}


\subsection{Ablation Study}
To minimize the influence of other components in the method, we choose CoOp+AnchorOPT as the baseline. For experiments on ImageNet, we set the anchor length to 1 and the soft token length to 6 by default. 
Complete experimental results are attached in the Appendix.

\noindent\textbf{Ensemble Operation.}
In this work, we average predictions from two prompts as the final output for novel classes. Table~\ref{table:ensemble} compares the generalization performance of AnchorOPT's learnable prompts against ATPrompt excluding anchor prompts. Our normal prompts outperform ATPrompt even without anchor integration, demonstrating superior adaptability.

\begin{table}[ht]
    \centering
    \resizebox{0.83\linewidth}{!}
    {
        \begin{tabular}{ccccc}
        \toprule
        Method   & Base  & Novel & HM    \\
        \midrule
        ATPrompt & 82.68 & 68.04 & 74.65 \\
        AnchorOPT w/o Ensemble & 81.24 & 72.46 & \textbf{76.60}  \\
        \bottomrule
        \end{tabular}
    }
    \vspace{-5pt}
    \caption{Comparison of learnable normal prompt performance on 11 datasets. Even without the integration of anchor prompts, our method still outperforms previous work. }
    \label{table:ensemble}
    \vspace{-5pt}
\end{table}

\noindent\textbf{Preposition ``of".} 
We employ the preposition "of" to construct anchor prompts because it can establish a possessive relationship between anchor and class tokens, aligning with our training objectives for anchor tokens. Table~\ref{table:preposition} evaluates preposition impacts on ImageNet, showing that irrelevant prepositions significantly impair novel-class generalization by failing to guide anchor token training effectively.



\begin{table}[h]
    \centering
    \resizebox{0.82\linewidth}{!}
    {
        \begin{tabular}{ccccc}
        \toprule
        Preposition Type & Value   & Base  & Novel & HM    \\
        \midrule
        \multirow{2}[2]{*}{Similar} & ``with" & 75.96 & 71.49 & 73.66 \\
        ~ & ``at"   & 76.07 & 71.54 & 73.74 \\
        \midrule
        \multirow{2}*{Irrelevant} & ``sun" & 75.96 & 71.05 & 73.42 \\
        ~ & ``sea" & 75.69 & 70.89 & 73.21 \\
        \midrule
        Ours & ``of" & 76.30 & 71.56 & \textbf{73.85} \\
        \bottomrule
        \end{tabular}
    }
    \vspace{-5pt}
    \caption{Comparison of different preposition words on ImageNet. The preposition ``of" works best.}
    \label{table:preposition}
    \vspace{-5pt}
\end{table}



\noindent\textbf{Anchor Length.}
With fixed soft token length~(6), Table~\ref{table:anchor_length} shows that anchor length critically influences performance. An anchor length of 1 yields optimal results, while longer anchors progressively degrade performance due to excessive constraints on soft token adaptability. Notably, while ATPrompt identifies 2–3 explicit attribute anchors as optimal, our method achieves peak performance with a single implicit anchor, indicating greater representational efficiency and robustness in learned anchors.



\begin{table}[ht]
    \centering
    \resizebox{0.88\linewidth}{!}
    {
        \begin{tabular}{ccccccc}
        \toprule
        Type & \multicolumn{3}{c}{Anchor Prompt} & \multicolumn{3}{c}{Normal Prompt} \\
        \cmidrule(lr){1-1}\cmidrule(lr){2-4}\cmidrule(lr){5-7}
        Length & Base & Novel & HM & Base & Novel & HM    \\
        \midrule
        1 & 74.63 & 71.03 & 72.79 & 76.30 & 69.51 & 72.75 \\
        2 & 74.58 & 70.82 & 72.65 & 72.14 & 63.32 & 67.44 \\
        3 & 74.54 & 71.17 & 72.82 & 34.23 & 11.17 & 16.84 \\
        4 & 74.54 & 71.12 & 72.79 & 13.41 & 1.52  & 2.73  \\
        \bottomrule
        \end{tabular}
    }
    \caption{Comparison of different anchor lengths on ImageNet. Optimal performance is achieved when the anchor length is 1.}
    \label{table:anchor_length}
    \vspace{-10pt}
\end{table}



\noindent\textbf{Anchor Arrangement.} 
Table~\ref{table:anchor_arrangement} evaluates token arrangement strategies—adaptive versus fixed positioning—with anchor token length fixed at 1. The adaptive arrangement, dynamically adjusted via the position matrix, consistently outperforms fixed-position variants on ImageNet.


\begin{table}[ht]
    \centering
    \resizebox{0.84\linewidth}{!}
    {
        \begin{tabular}{ccccc}
        \toprule
        Type & Position & Base  & Novel & HM    \\
        \midrule
        Adaptive & Position Matrix & 76.30 & 71.56 & \textbf{73.85} \\
        \midrule
        \multirow{2}[3]{*}{Fixed} & Before Soft Token & 76.03 & 71.27 & 73.57 \\
                 & Middle            & 75.99 & 71.17 & 73.50 \\
                 & After Class Token & 76.23 & 71.31 & 73.69 \\
        \bottomrule
        \end{tabular}
    }
    \caption{Comparison of different anchor token arrangements on ImageNet. Using a position matrix yields the best results. }
    \label{table:anchor_arrangement}
    \vspace{-5pt}
\end{table}

\noindent\textbf{Distillation.} Table~\ref{table:distillation} quantifies the impact of distillation on generalization performance. Incorporating distillation from ensemble predictions enables soft tokens to acquire more generalizable representations for novel classes.


\begin{table}[ht]
    \centering
    \resizebox{0.48\linewidth}{!}
    {
        \begin{tabular}{cccc}
        \toprule
        KD   & Base  & Novel & HM    \\
        \midrule
        ~          & 75.86 & 71.36 & 73.54 \\
        \checkmark & 76.30 & 71.56 & 73.85 \\
        \bottomrule
        \end{tabular}
    }
    \caption{Ablation of the knowledge distillation strategy on ImageNet. Removing the distillation term weakens overall generalization performance.}
    \label{table:distillation}
    \vspace{-10pt}
\end{table}

\section{Conclusion}


We propose AnchorOPT, a dynamic anchor-based prompt learning framework that optimizes explicit fixed anchors into implicit learnable tokens and introduces a position matrix for adaptive prompt sequencing. Our method employs a two-stage training paradigm to sequentially optimize anchor tokens and downstream adaptation of soft prompts and position matrix. To ensure compatibility across architectures, we develop shallow and deep variants. Extensive experiments across 11 datasets demonstrate that AnchorOPT, combined with CoOp, achieves competitive performance against more complex methods, revealing significant untapped potential in foundational prompt learning templates. These results indicate that basic prompt structures, since CoOp's introduction, remain suboptimal and warrant further exploration.



{
    \small
    \bibliographystyle{ieeenat_fullname}
    \bibliography{main}
}


\newpage
\maketitlesupplementary

\setcounter{table}{0}
\setcounter{figure}{0}
\setcounter{section}{0}
\setcounter{equation}{0}

\renewcommand{\thesection}{S\arabic{section}}
\renewcommand{\thetable}{S\arabic{table}}
\renewcommand{\thefigure}{S\arabic{figure}}
\renewcommand{\theequation}{S\arabic{equation}}

\begin{figure*}[t]
    \centering
    \includegraphics[width=0.99\linewidth]{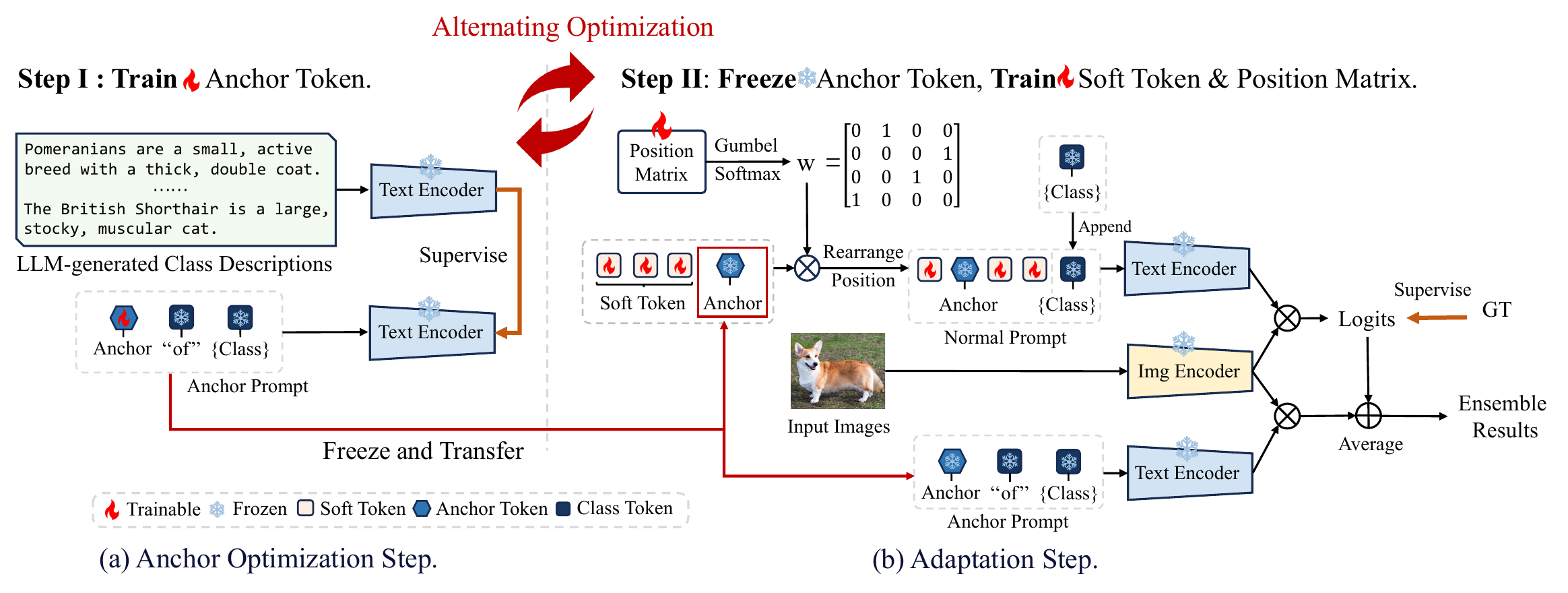}
    \caption{Illustration of one-stage training paradigm. The framework alternates between (i) optimizing anchor tokens and (ii) updating soft tokens and the position matrix while freezing anchors, iterating until convergence.}
    \label{fig:one-stage}
\end{figure*}

\section{One-stage Training Paradigm Extension}
Fig.~\ref{fig:one-stage} illustrates the one-stage training paradigm, which unifies the previous two-stage process into sequential optimization steps within a single framework. This approach alternates between optimizing target parameters while freezing non-target parameters, yielding stable convergence. By eliminating stage separation, the paradigm reduces training complexity while maintaining performance integrity. 
Table~\ref{table:anchoropt_all_accuracy} shows the results of the one-stage training method on 11 datasets.
Our method adopts the two-stage paradigm, which demonstrates two key advantages over one-stage alternatives: (1) enhanced classification accuracy, and (2) streamlined hyperparameter optimization through elimination of redundant anchor token retraining during each training session.



\section{Implementation Details}

\subsection{Dataset}
We evaluate the performance of our method on 15 recognition datasets. For generalization from base-to-novel classes and cross-dataset evaluation, we evaluate the performance of our method on 11 diverse recognition datasets. Specifically, these datasets include ImageNet-1K~\cite{deng2009imagenet} and Caltech-101~\cite{fei2004learning} for generic object classification; OxfordPets~\cite{parkhi2012cats}, Stanford Cars~\cite{krause20133d}, Flowers102~\cite{nilsback2008automated}, Food101~\cite{bossard2014food}, and FGVC Aircraft~\cite{maji2013fine} for fine-grained classification, SUN-397~\cite{xiao2010sun} for scene recognition, UCF-101~\cite{soomro2012ucf101} for action recognition, DTD~\cite{cimpoi2014describing} for texture classification, and EuroSAT~\cite{helber2019eurosat} for satellite imagery recognition. For domain generalization experiments, we use ImageNet-1K~\cite{deng2009imagenet} as the source dataset and its four variants as target datasets, including ImageNet-V2~\cite{recht2019imagenet}, ImageNet-Sketch~\cite{wang2019learning}, ImageNet-A~\cite{hendrycks2021natural}, and ImageNet-R~\cite{hendrycks2021many}.

\begin{algorithm}[t]
    \caption{Pseudocode of AnchorOPT in PyTorch.}
    \label{algo:anchoropt_algo}
    \footnotesize
    \vspace{-10pt}
    \begin{alltt}\color{ForestGreen}

# anc_token: anchor token
# soft_token: soft token
# cls_token: class token
# N: length of anchor tokens
# M: length of soft tokens
# t_anc: anchor prompt
# pos_mat: position matrix
# f_t: text encoder
# f_v: image encoder

\color{ForestGreen}# stage 1: anchor optimization \color{Black}
for des_text in description_dataset:
    feat_des = f_t(des_text)
    feat_anc = f_t(t_anc)
    loss = MSE(feat_anc, feat_des)
    loss.backward()
    
\color{ForestGreen}# stage 2: adaptation\color{Black}
\color{ForestGreen}# calculate position matrix\color{Black}
pos_params = torch.randn(N+M, N+M)
pos_params = nn.Parameter(pos_params,
             requires_grad=True)
pos_mat = F.gumbel_softmax(pos_params, hard=True)

for img, label in labeled_dataset:
    feat_img = f_v(img)
    t_norm = [soft_token, anc_token] * pos_mat
    feat_norm = f_t([t_norm, cls_token])

    \color{ForestGreen}# predictions\color{Black}
    l_anc = feat_img * feat_anc.t()
    l_norm = feat_img * feat_norm.t()
    l_ens = (l_anc + l_norm)/2

    \color{ForestGreen}# calculate loss \color{Black}
    loss_kd = KL(l_norm, l_ens)
    loss_ce = CE(l_norm, label)
    loss = alpha * loss_ce + beta * loss_kd
    loss.backward()
    
\end{alltt}
\vspace{-10pt}
\end{algorithm}

\subsection{Training Details}
All experiments were conducted on a single NVIDIA H20 GPU. We use the ViT-B/16 CLIP as our default model. We adopted the standard data augmentation scheme as the baseline method, including random resized cropping and flipping. We employed Stochastic Gradient Descent~(SGD) as the optimizer for both soft tokens, anchor tokens, and the position matrix. The length of the anchor token is set to 1. We report base and novel class accuracy and their harmonic mean~(HM) averaged over 3 runs. Algorithm~\ref{algo:anchoropt_algo} provides AnchorOPT's PyTorch-style pseudocode.

\subsubsection{Category Description Generation}

Inspired by previous works~\cite{pratt2023does}, we leverage a Large Language Model~(LLM) such as GPT-3\cite{brown2020language} to generate class-specific descriptions through the query template: ``\textit{What does a \{Class\} look like?}'' The resulting LLM-generated descriptions are encoded into text features via a text encoder, which serves as the training objective for anchor prompts. In Table~\ref{table:description}, we show a portion of the category descriptions generated based on the category name on the Caltech101 dataset.

\begin{table*}[t]
    \begin{subtable}{\textwidth}
    \centering
    \resizebox{0.96\linewidth}{!}
    {
    \centering
    \begin{tabular}{cccc|ccc|ccc|ccc}
    \toprule 
    \multirow{2}[3]{*}{Method} & \multicolumn{3}{c}{Average} & \multicolumn{3}{c}{ImageNet} & \multicolumn{3}{c}{ Caltech101} & \multicolumn{3}{c}{OxfordPets} \\
    \cmidrule(lr){2-4}\cmidrule(lr){5-7}\cmidrule(lr){8-10}\cmidrule(lr){11-13}
    & Base & Novel & HM & Base & Novel & HM & Base & Novel & HM & Base & Novel & HM \\
    \midrule
    CoOp~\scriptsize{\textcolor{gray}{(IJCV 22)}} & 82.69 & 63.22 & 71.66 & 76.47 & 67.88 & 71.92 & 98.00 & 89.81 & 93.73 & 93.67 & 95.29 & 94.47 \\
    ATPrompt~\scriptsize{\textcolor{gray}{(ICCV 25)}} & 82.68 & 68.04 & 74.65  & 76.27 & 70.60 & 73.33 & 97.95 & 93.63 & 95.74 & 94.77 & 96.59 & 95.67 \\ 
    \midrule
    AnchorOPT-One Stage & 80.36 & 75.89 & 78.06 & 75.37 & 71.41 & 73.34 & 97.67 & 95.38 & 96.51 & 94.72 & 98.04 & 96.35 \\
    \midrule
    AnchorOPT-Anchor Prompts & 71.69 & 76.93 & 73.91 & 74.63 & 71.03 & 72.78 & 97.93 & 95.42 & 96.53 & 94.31 & 98.15 & 96.19 \\
    AnchorOPT-Normal Prompts & \textbf{81.24} & 72.46 & 76.60 & \textbf{76.30} & 69.51 & 72.75 & \textbf{98.26} & 95.12 & 96.66 & \textbf{95.32} & 97.02 & 96.16 \\
    AnchorPOT-Ensemble & 79.72 & \textbf{76.27} & 77.96 & 76.36 & \textbf{71.56} & 73.88 & 98.28 & \textbf{95.34} & 96.79 & 95.66 & \textbf{98.04} & 96.84 \\
    \midrule
    AnchorPOT-Final & \textbf{81.24} & \textbf{76.27} & \textbf{78.68} & \textbf{76.30} & \textbf{71.56} & \textbf{73.85} & \textbf{98.26} & \textbf{95.34} & \textbf{96.78} & \textbf{95.32} & \textbf{98.04} & \textbf{96.66}  \\
    \bottomrule
    \end{tabular}
    }
    \end{subtable}

    \vspace{5pt}
    \begin{subtable}{\textwidth}
    \centering
    \resizebox{0.96\linewidth}{!}
    {
    \begin{tabular}{cccc|ccc|ccc|ccc}
    \toprule \multirow{2}[3]{*}{ Method } & \multicolumn{3}{c}{StanfordCars} & \multicolumn{3}{c}{ Flowers102} & \multicolumn{3}{c}{ Food101} & \multicolumn{3}{c}{ FGVCAircraft} \\
    \cmidrule(lr){2-4}\cmidrule(lr){5-7}\cmidrule(lr){8-10}\cmidrule(lr){11-13} 
    & Base & Novel & HM & Base & Novel & HM & Base & Novel & HM & Base & Novel & HM \\
    \midrule
    CoOp~\scriptsize{\textcolor{gray}{(IJCV 22)}} & 78.12 & 60.40 & 68.13 & 97.60 & 59.67 & 74.06 & 88.33 & 82.26 & 85.19 & 40.44 & 22.30 & 28.75 \\
    ATPrompt~\scriptsize{\textcolor{gray}{(ICCV 25)}} & 77.43 & 66.55 & 71.58 & 97.44 & 67.52 & 79.77 & 88.74 & 87.44 & 88.09 & 40.38 & 27.22 & 32.52 \\
    \midrule
    AnchorOPT-One Stage & 75.43 & 74.15 & 74.78 & 96.33 & 75.55 & 84.68 & 90.12 & 91.93 & 91.02 & 35.95 & 35.15 & 35.55 \\
    \midrule
    AnchorOPT-Anchor Prompts & 64.57 & 75.27 & 69.50 & 73.41 & 78.86 & 76.04 & 90.04 & 91.62 & 90.82 & 28.87 & 36.65 & 32.30 \\
    AnchorOPT-Normal Prompts & \textbf{76.73} & 69.68 & 73.04 & \textbf{97.22} & 68.13 & 80.12 & \textbf{90.40} & 91.23 & 90.81 & \textbf{38.48} & 29.42 & 33.35 \\
    AnchorOPT-Ensemble & 74.80 & \textbf{74.43} & 74.61 & 92.78 & \textbf{75.41} & 84.94 & 90.64 & \textbf{91.95} & 91.29 & 36.98 & \textbf{35.47} & 36.21 \\
    \midrule
    AnchorOPT-Final & \textbf{76.73} & \textbf{74.43} & \textbf{75.67} & \textbf{97.22} & \textbf{75.41} & \textbf{84.94} & \textbf{90.40} & \textbf{91.95} & \textbf{91.17} & \textbf{38.48} & \textbf{35.47} & \textbf{36.91} \\
    \bottomrule
    \end{tabular}
    }
    \end{subtable}

    \vspace{5pt}
    \begin{subtable}{\textwidth}
    \centering
    \resizebox{0.96\linewidth}{!}
    {
    \begin{tabular}{cccc|ccc|ccc|ccc}
    \toprule 
    \multirow{2}[3]{*}{Method} & \multicolumn{3}{c}{SUN397} & \multicolumn{3}{c}{DTD} & \multicolumn{3}{c}{EuroSAT} & \multicolumn{3}{c}{UCF101} \\
    \cmidrule(lr){2-4}\cmidrule(lr){5-7}\cmidrule(lr){8-10}\cmidrule(lr){11-13} 
    & Base & Novel & HM & Base & Novel & HM & Base & Novel & HM & Base & Novel & HM \\
    \midrule
    CoOp~\scriptsize{\textcolor{gray}{(IJCV 22)}} & 80.60 & 65.89 & 72.51 & 79.44 & 41.18 & 54.24 & 92.19 & 54.74 & 68.69 & 84.69 & 56.05 & 67.46 \\
    ATPrompt~\scriptsize{\textcolor{gray}{(ICCV 25)}} & 80.84 & 68.64 & 74.24 & 80.83 & 45.49 & 58.22 & 90.34 & 59.79 & 71.96 & 84.49 & 64.96 & 73.45 \\
    \midrule
    AnchorOPT-One Stage & 80.62 & 78.19 & 79.39 & 76.28 & 59.62 & 66.93 & 79.35 & 77.18 & 78.25 & 82.13 & 78.24 & 80.14 \\
    \midrule
    AnchorOPT-Anchor Prompts & 74.36 & 78.37 & 76.31 & 55.78 & 62.56 & 58.98 & 57.33 & 80.20 & 66.86 & 75.33 & 78.15 & 76.71 \\
    AnchorOPT-Normal Prompts & \textbf{79.67} & 75.91 & 77.74 & \textbf{74.04} & 57.55 & 64.76 & \textbf{83.75} & 66.90 & 74.38 & \textbf{83.45} & 76.56 & 79.86 \\
    AnchorOPT-Ensemble & 79.42 & \textbf{78.73} & 79.07 & 72.11 & \textbf{61.72} & 66.51 & 77.34 & \textbf{77.14} & 77.24 & 82.50 & \textbf{79.18} & 80.81 \\
    \midrule
    AnchorOPT-Final & \textbf{79.67} & \textbf{78.73} & \textbf{79.20} & \textbf{74.04} & \textbf{61.72} & \textbf{67.32} & \textbf{83.75} & \textbf{77.14} & \textbf{80.31} & \textbf{83.45} & \textbf{79.18} & \textbf{81.26} \\
    \bottomrule
    \end{tabular}
    }
    \end{subtable}
    \caption{Accuracy of each prompt in AnchorOPT.}
    \label{table:anchoropt_all_accuracy}
\end{table*}

\begin{figure*}[ht]
    \centering
    \includegraphics[width=0.65\linewidth]{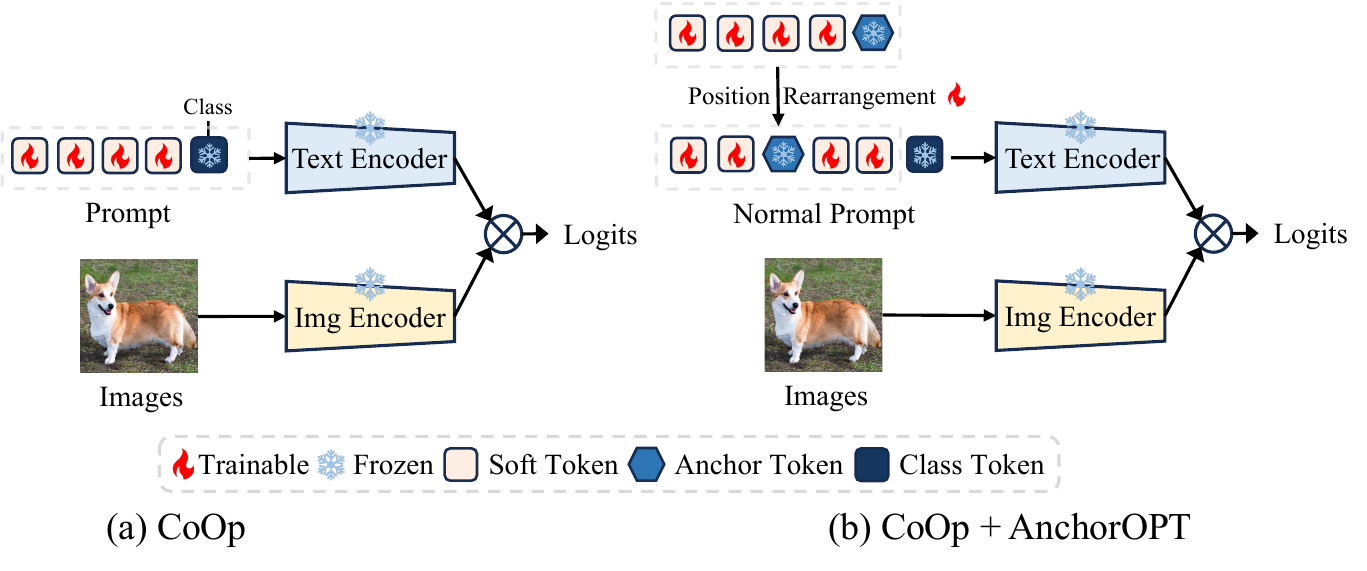}
    \caption{Architecture comparison between CoOp and CoOp+AnchorOPT. In CoOp+AnchorOPT, anchor terms are first pre-trained, then embedded into normal prompts, and multiplied with the position matrix. The final result is used as the input to the encoder.}
    \label{fig:dyntoken+coop}
\end{figure*}

\begin{figure*}[ht]
    \centering
    \includegraphics[width=0.99\linewidth]{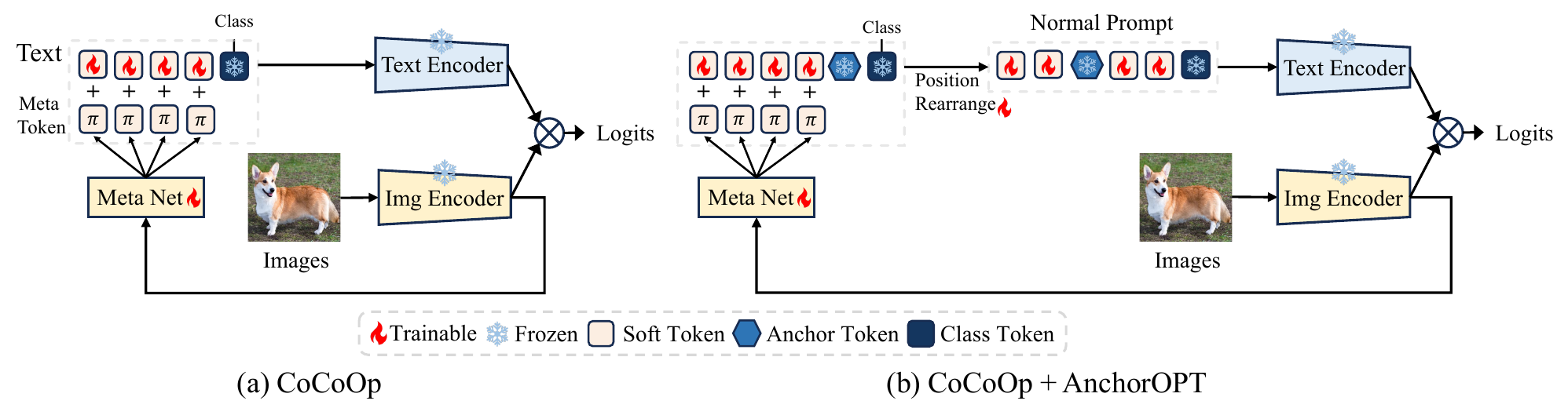}
    \caption{Architecture comparison between CoCoOp and CoCoOp+AnchorOPT. In CoCoOp+AnchorOPT, soft tokens are first augmented by the offset generated by the meta-network and subsequently multiplied by the position matrix. }
    \label{fig:dyntoken+cocoop}
\end{figure*}

\begin{figure*}[ht]
    \centering
    \includegraphics[width=0.95\linewidth]{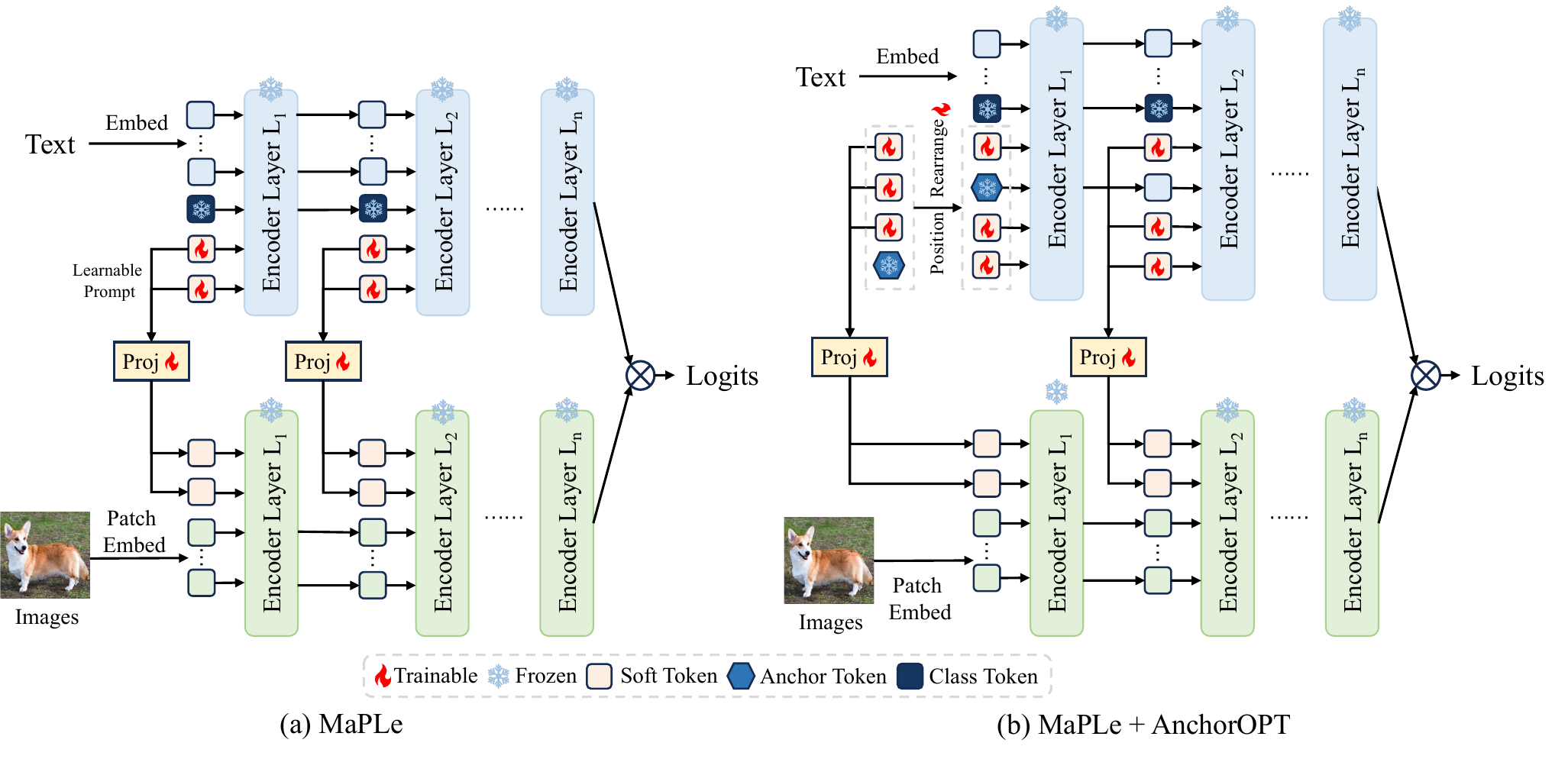}
    \caption{Architecture comparison between MaPLe and MaPLe+AnchorOPT. In MaPLe+AnchorOPT, the initial soft textual tokens are first mapped to visual tokens, which are subsequently multiplied by the position matrix. During forward propagation, the representations derived from the anchor tokens are preserved to provide semantic guidance.}
    \label{fig:dyntoken+maple}
\end{figure*}

\subsubsection{CoOp+AnchorOPT} 
A batch size of 8 and an initial learning rate of 0.0005 are employed. Models train for 100 epochs on Caltech-101, Oxford Pets, Stanford Cars, Flowers-102, FGVC Aircraft, DTD, EuroSAT, and UCF-101; 20 epochs on ImageNet, Food-101, and SUN-397. Soft token lengths are set to 4 for Caltech-101, Oxford Pets, Flowers-102, Food-101, FGVC Aircraft, DTD, and EuroSAT; 6 for ImageNet and UCF-101; and 8 for SUN-397. Anchor token length remains fixed at 1 across all datasets. Architectural integration of AnchorOPT into CoOp is illustrated in Fig.~\ref{fig:dyntoken+coop}.

\subsubsection{CoCoOp+AnchorOPT}
Consistent with the baseline, a batch size of 1 and an initial learning rate of 0.002 are used. Training spans 20 epochs for Caltech-101, Oxford Pets, Stanford Cars, Flowers-102, Food-101, FGVC Aircraft, DTD, EuroSAT, and UCF-101; 10 epochs for SUN-397 and ImageNet. Soft token lengths are set to 4 for all datasets except SUN-397 (8). Anchor token length remains uniformly 1. Integration details appear in Fig.~\ref{fig:dyntoken+cocoop}. Soft tokens are processed by meta-network-generated offsets, then multiplied by the position matrix to form text encoder inputs.

\subsubsection{MaPLe+AnchorOPT}
\begin{figure*}[h]
    \centering
    \includegraphics[width=0.58\linewidth]{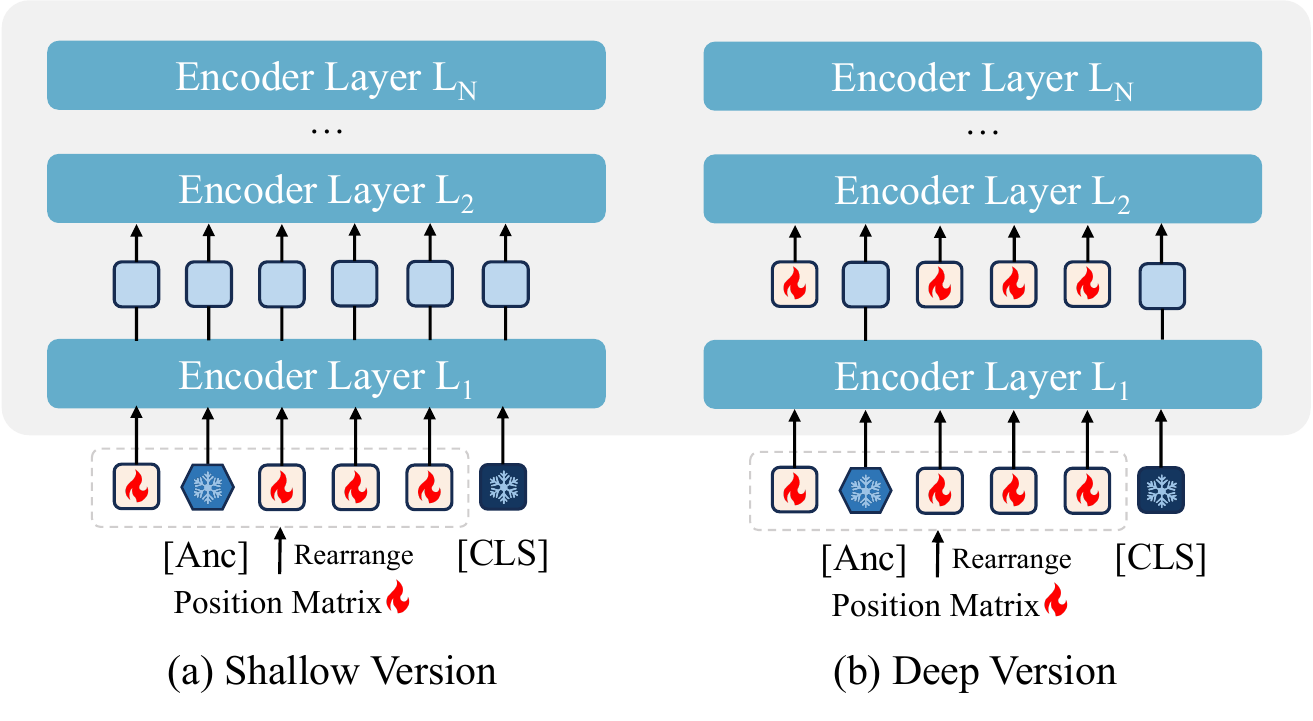}
    \caption{Architecture comparison between shallow version and deep version. In the deep version, the representations derived from the anchor tokens are preserved to provide semantic guidance.}
    \label{fig:shallow_and_deep}
\end{figure*}

With a batch size of 4 and an initial learning rate of 0.0035, soft token lengths are set to 4 for Stanford Cars, Oxford Pets, FGVC Aircraft, SUN-397, DTD, and UCF-101; and 8 for Caltech-101, Oxford Pets, Food-101, and EuroSAT. Anchor token length remains fixed at 1. Integration is shown in Fig.~\ref{fig:dyntoken+maple}. To preserve token diversity amid position-matrix operations (which involve selective discarding and repetition), initial text soft tokens are mapped to visual tokens before position-matrix multiplication. The resulting tokens serve as text encoder inputs, with deeper-layer soft tokens undergoing position-assigned discarding and reintroduction. The detailed comparison between the shallow and deep versions is presented in Fig.~\ref{fig:shallow_and_deep}.

\begin{figure*}[ht]
    \centering
    \includegraphics[width=0.72\linewidth]{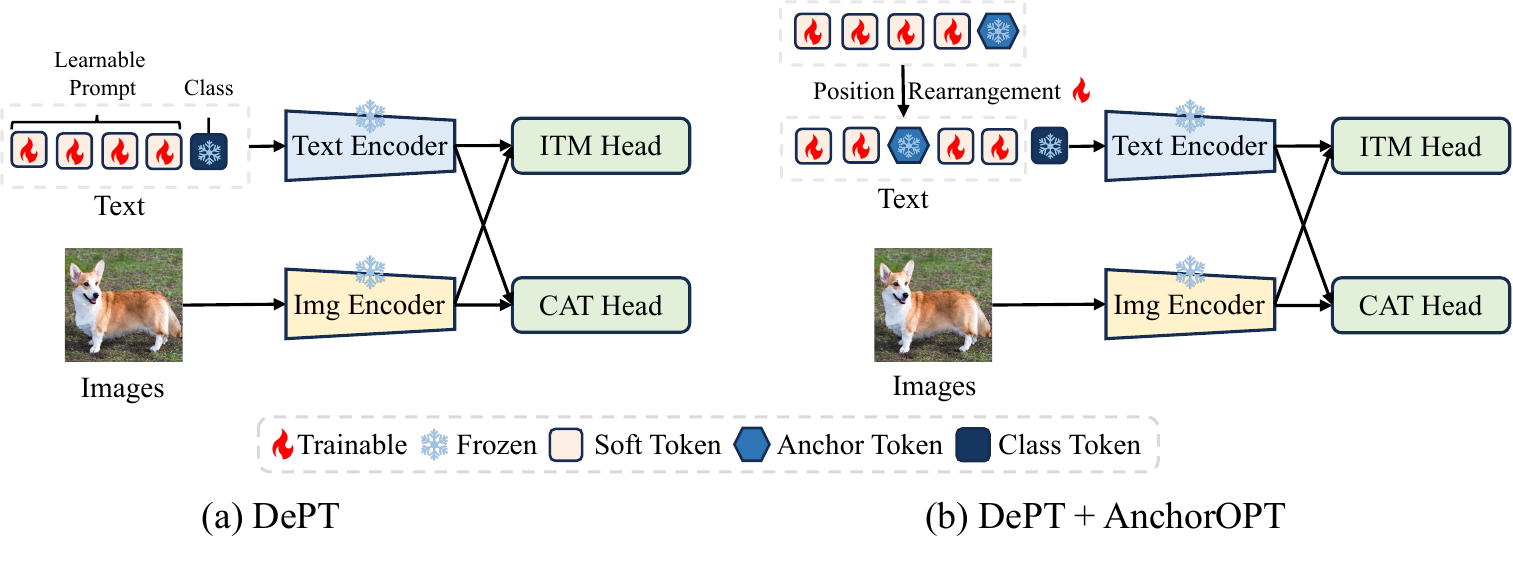}
    \caption{Architecture comparison between DePT and DePT+AnchorOPT. DePT+AnchorOPT uses the same operations as CoOp+AnchotOPT to construct the input prompt portion.}
    \label{fig:dyntoken+dept}
\end{figure*}

\subsubsection{DePT+AnchorOPT}
Using a batch size of 4 and an initial learning rate of 0.0035, soft token lengths are set to 4 for Caltech-101, Food-101, FGVC Aircraft, and ImageNet; and 8 for Oxford Pets, Stanford Cars, Flowers-102, SUN-397, DTD, EuroSAT, and UCF-101. Anchor token length remains consistently 1. Integration details are provided in Fig.~\ref{fig:dyntoken+dept}. Final novel-class predictions are derived by ensembling anchor prompt outputs with results from the image-text matching~(ITM) head.


\section{Experimental Results}

\subsection{Base-to-Novel Experiments}

\textbf{Accuracy of Each Prompt.}
Table~\ref{table:anchoropt_all_accuracy} reports the accuracy of the selected anchor prompt, normal prompt, ensemble prediction, and the final results.
In this work, we integrate anchor and normal prompt predictions via equal-weighted averaging $q_{ens}=(q_{norm}+q_{anc})/2$. Equal weighting is adopted because it eliminates hyperparameter tuning while treating both prediction sources as equally reliable—a critical simplification for real-world deployment where computational efficiency and robustness are prioritized.

\subsection{Ablation Study}

\textbf{Preposition ``None".}
Table~\ref{table:preposition_none} presents results for anchor prompts omitting prepositions. This configuration avoids performance degradation from irrelevant tokens (e.g., "sun" or "sea"), maintaining strong generalization—though marginally inferior to preposition-augmented prompts.

\begin{table}[h]
    \centering
    \resizebox{0.84\linewidth}{!}
    {
        \begin{tabular}{ccccc}
        \toprule
        Preposition Type & Value   & Base  & Novel & HM    \\
        \midrule
        None & None   & 75.87 & 71.43 & 73.58 \\
        \midrule
        \multirow{2}*{Similar} & ``with" & 75.96 & 71.49 & 73.66 \\
        ~ & ``at"   & 76.07 & 71.54 & 73.74 \\
        \midrule
        \multirow{2}*{Irrelevant} & ``sun" & 75.96 & 71.05 & 73.42 \\
        ~ & ``sea" & 75.69 & 70.89 & 73.21 \\
        \midrule
        Ours & ``of" & 76.30 & 71.56 & \textbf{73.85} \\
        \bottomrule
        \end{tabular}
    }
    \caption{Comparison of different preposition words on ImageNet. The absence of prepositions affects the model's ability to generalize to base classes.}
    \label{table:preposition_none}
\end{table}

\noindent\textbf{Gumbel Softmax Temperature.}
While we fix the Gumbel Softmax temperature at 1 during primary experiments, Table~\ref{table:temperature_gumbel_softmax} demonstrates that values spanning 0.1–4 yield comparable performance. This indicates model robustness to temperature variations in the Gumbel Softmax function.

\begin{table}[ht]
    \centering
    \resizebox{0.53\linewidth}{!}
    {
        \begin{tabular}{cccc}
        \toprule
        Temp & Base  & Novel & HM   \\
        \midrule
        0.1 & 76.08 & 71.34 & 73.63 \\
        0.5 & 76.10 & 71.39 & 73.67 \\
        1 & 76.30 & 71.56 & \textbf{73.85} \\
        2 & 75.94 & 71.48 & 73.64 \\
        4 & 75.94 & 71.58 & 73.70 \\
        \bottomrule
        \end{tabular}
    }
    \caption{Comparison of different temperature values of Gumbel Softmax on ImageNet. The varying temperature values have little impact on the final model performance, and the fluctuations are all within the normal range.}
    \label{table:temperature_gumbel_softmax}
\end{table}

\section{Discussion}

\textbf{Visualization of Learned Anchor Tokens.}
To visualize learned anchor embeddings, we project embeddings into the token vocabulary space via inverse embedding. However, the resulting sequences consist of incoherent character strings that form neither valid words nor semantically meaningful content. This limitation stems from the fundamental incompatibility between the continuous latent space in which anchor tokens are optimized and the discrete tokenization process of CLIP, preventing meaningful mapping from continuous embeddings to human-interpretable text.

\noindent\textbf{Exclusion of Class Tokens During Position Rearrangement.} The class token contains Class tokens that encode task-critical semantic information essential for representation fidelity. Integrating them into position rearrangement introduces two significant challenges: (1) displacement or omission of class tokens compromises prompt efficacy and disrupts semantic representation learning; (2) increased positional flexibility elevates optimization complexity for soft prompts and anchors, degrading model performance.

\section{Future Work}
Current prompt learning frameworks remain predominantly evaluated on CLIP, with limited adaptation to alternative vision-language models (VLMs) such as SigLIP and EVA-CLIP. We will address this gap by developing a cross-model benchmark to systematically evaluate and enhance prompt learning compatibility across diverse VLM architectures.

\begin{table*}[ht]
    \centering
    \resizebox{0.95\linewidth}{!}
    {
        \begin{tabular}{cc}
        \toprule
        Class & Description \\
        \midrule
        \multirow{5}*{gerenuk} & A gerenuk is a type of antelope that is native to Africa and Asia. \\
        ~ & A gerenuk is a lanky, long-necked gazelle with narrow, somewhat pointed horns that curve downward. \\
        ~ & A gerenuk is a long-necked antelope with thin legs and large ears. \\
        ~ & A gerenuk is a tall, slender antelope with long, thin legs. \\
        ~ & A gerenuk is a type of antelope with a long neck and limbs. \\
        \midrule
        \multirow{5}*{background} & A background looks like an image or color behind the main focus of a piece.", \\
        ~ & A background looks like the area behind the main subject of an image. \\
        ~ & A background usually has a certain color or pattern. \\
        ~ & A background is an image or color that appears behind the main content on a screen. \\
        ~ & A background generally refers to the parts of an image that are farthest from the viewer. \\
        \midrule
        \multirow{5}*{hawksbill} & A hawksbill sea turtle has a narrow head with a hawk-like beak. \\
        ~ & A hawksbill sea turtle has a hawk-like beak, which is where it gets its name. \\
        ~ & A hawksbill looks like a small, thin turtle with a long neck and a beak-like mouth. \\
        ~ & A hawksbill looks like a small turtle with a long, pointed beak. \\
        ~ & A hawksbill sea turtle is a small to medium-sized turtle that can grow up to about 3 feet long. \\
        \midrule
        \multirow{5}*{headphone} & A headphone typically consists of two small speakers that are attached to a headband. \\
        ~ & A headphone is a small speaker that you can wear on your head. \\
        ~ & A headphone is a small pair of speakers that fit over a person's ears.\\
        ~ & A headphone typically consists of a pair of small speakers that are placed over the ears. \\
        ~ & Most headphones are designed to go over your ears. \\
        \midrule
        \multirow{5}*{ant} & A ant is small, black, and has six legs. \\
        ~ & The ant is a small, black creature with six legs. \\
        ~ & A ant is a small, black insect with six legs. \\
        ~ & A typical ant is about 2.  \\
        ~ & A ant typically has a dark brown or black body with a narrow waist. \\
        \midrule
        \multirow{5}*{butterfly} & Most butterflies have brightly colored wings with intricate patterns. \\
        ~ & A butterfly typically has two large wings that are decorated with brightly colored scales. \\
        ~ & Most butterflies have brightly colored wings, with patterns made up of tiny scales. \\
        ~ & A butterfly is a flying insect. \\
        ~ & A butterfly has brightly colored wings and a long thin body. \\
        \midrule
        \multirow{5}*{lamp} & A lamp is a household appliance that is used to illuminate a room. \\
        ~ & When most people think of a lamp, they think of a light bulb on a stand with a shade. \\
        ~ & A lamp is a household appliance that consists of a light bulb inside a housing, which is usually made of glass. \\
        ~ & A lamp looks like a small lightbulb on a metal stand. \\
        ~ & A lamp is a device that contains an electric light bulb and is used to produce light. \\
        \midrule
        \multirow{5}*{strawberry} & A strawberry is a fruit that is generally red, although some varieties are yellow, white, or light green. \\
        ~ & A strawberry is a small, red, spherical fruit with a seed-studded surface. \\
        ~ & A strawberry is a small red fruit with a seed-covered surface. \\
        ~ & A strawberry is a small, red fruit that is covered in tiny seeds. \\
        ~ & A strawberry typically has a bright red exterior with small seeds on the surface. \\
        \bottomrule
        \end{tabular}
    }
    \caption{A partial display of the descriptions generated from the category names on the Caltech101 dataset.}
    \label{table:description}
\end{table*}
\end{document}